\documentclass[10pt,twocolumn,letterpaper]{article}

\usepackage[pagenumbers]{cvpr} 

\definecolor{cvprblue}{rgb}{0.21,0.49,0.74}
\usepackage[pagebackref,breaklinks,colorlinks,allcolors=cvprblue]{hyperref}

\usepackage[utf8]{inputenc}
\usepackage[T1]{fontenc}
\usepackage{hyperref}
\usepackage{url}
\usepackage{booktabs}
\usepackage{amsfonts}
\usepackage{nicefrac}
\usepackage{microtype}
\usepackage[dvipsnames]{xcolor}
\usepackage[many]{tcolorbox}
\usepackage{amssymb}
\usepackage{colortbl}
\usepackage{pifont}
\usepackage{multirow}
\usepackage{graphicx}
\usepackage{tabularx}
\usepackage{threeparttable}
\usepackage{wrapfig}
\usepackage{enumitem}
\usepackage{subcaption}
\usepackage{xspace}
\usepackage{ragged2e}
\usepackage{gradient-text}

\newcommand{\cmark}{\ding{51}}
\newcommand{\xmark}{\ding{55}}

\newcommand{\model}{ResAgent\xspace}

\def\paperID{*****} 
\def\confName{CVPR}
\def\confYear{2026}

\title{ ResAgent: Entropy-based Prior Point Discovery and Visual Reasoning \\ for Referring Expression Segmentation }

\author{
Yihao Wang\textsuperscript{1} \quad
Jusheng Zhang\textsuperscript{1} \quad
Ziyi Tang\textsuperscript{1} \quad
Keze Wang\textsuperscript{1} \quad
Meng Yang\textsuperscript{1} \\
\textsuperscript{1}Sun Yat-sen University \\
{\tt\small \{wangyh357, tangzy27\}@mail2.sysu.edu.cn, zhangjusheng19981128@gmail.com,}\\
{\tt\small kezewang@gmail.com, yangm6@mail.sysu.edu.cn}
}

\begin{document}
\maketitle

\begin{abstract}
Referring Expression Segmentation (RES) is a core vision-language segmentation task that enables pixel-level understanding of targets via free-form linguistic expressions, supporting critical applications such as human-robot interaction and augmented reality. Despite the progress of Multimodal Large Language Model (MLLM)-based approaches, existing RES methods still suffer from two key limitations: first, the coarse bounding boxes from MLLMs lead to redundant or non-discriminative point prompts; second, the prevalent reliance on textual coordinate reasoning is unreliable, as it fails to distinguish targets from visually similar distractors. To address these issues, we propose \textbf{\model}, a novel RES framework integrating \textbf{E}ntropy-\textbf{B}ased Point \textbf{D}iscovery (\textbf{EBD}) and \textbf{V}ision-\textbf{B}ased \textbf{R}easoning (\textbf{VBR}). Specifically, EBD identifies high-information candidate points by modeling spatial uncertainty within coarse bounding boxes, treating point selection as an information maximization process. VBR verifies point correctness through joint visual-semantic alignment, abandoning text-only coordinate inference for more robust validation. Built on these components, \model implements a coarse-to-fine workflow: bounding box initialization, entropy-guided point discovery, vision-based validation, and mask decoding. Extensive evaluations on four benchmark datasets (RefCOCO, RefCOCO+, RefCOCOg, and ReasonSeg) demonstrate that \model achieves new state-of-the-art performance across all four benchmarks, highlighting its effectiveness in generating accurate and semantically grounded segmentation masks with minimal prompts.

\end{abstract}
\section{Introduction}
\label{sec:intro}

\textbf{R}eferring \textbf{E}xpression \textbf{S}egmentation (RES) aims to bridge visual perception and natural language understanding by generating segmentation masks for objects referenced by free-form linguistic expressions \cite{kazemzadeh-etal-2014-referitgame,hu2016Segmentation,yu2016Modeling,Mao_2016_CVPR}. As a core vision--language grounding task, RES supports key applications such as human--robot interaction~\cite{xu2025Mobility,hannus2025iavla,liang2025pixelvla}, augmented reality~\cite{Gao2022VRbased,Serrano2017Movie,Wyssenbach2025Segmentation}, and content-based retrieval~\cite{An2023Towards,jiang2021Learning,wang2024fine}, where models must interpret fine-grained textual cues and precisely localize targets under cluttered or ambiguous visual contexts. 

Traditional RES approaches based on convolutional architectures \cite{chen2019see,feng2021encoder,huang2020referring,liu2017recurrent} or early cross-modal attention mechanisms \cite{ding2021vision,ding2022vlt,kim2022restr,liu2023caris,liu2023multi} often struggle to model the rich, fine-grained semantic alignment needed for robust grounding. With the rise of Multimodal Large Language Models (MLLMs), recent RES methods \cite{wu2024toward,xia2024gsva,lai2024lisa,Rasheed2024glamm,liu2025unipixel,Chen2025sam4mllm,lan2025textseg,zhu2025segAgent} leverage MLLMs to guide segmentation models such as SAM \cite{kirillov2023segment,ravi2024sam}, typically through coarse bounding box prediction or point-level prompting, thereby improving high-level reasoning and linguistic understanding.

\begin{figure}[t]
  \centering
   \includegraphics[width=\linewidth]{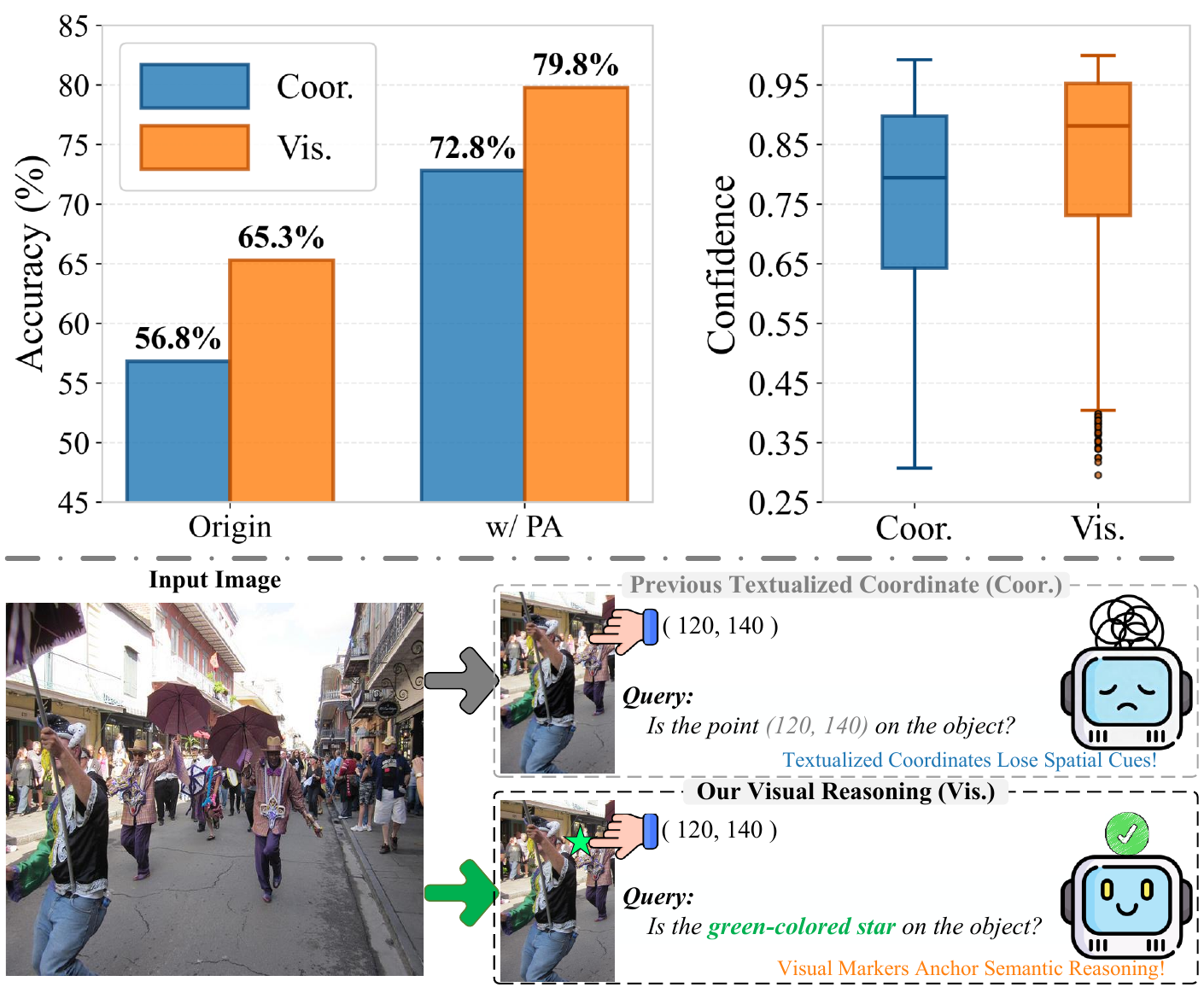}

   \caption{Top: Accuracy (left) and confidence (right) of coordinate-based versus visual reasoning for point inference in a zero-shot setting. Bottom: Example queries highlighting the fundamental difference—textual coordinates lose spatial cues, while visual reasoning provides semantic anchoring.
}
   \label{fig:intro}
\end{figure}



However, current MLLM-based RES pipelines still face two central limitations. \textit{First}, bounding boxes predicted by MLLMs are coarse and offer only approximate spatial localization, requiring point prompts for fine-grained segmentation; yet widely used point sampling strategies (e.g., random or uniform sampling) provide redundant or non-discriminative cues. \textit{Second}, MLLMs treat textualized coordinates as ordinary language tokens (e.g., ``(175, 60)''→``1–7–5–6–0''), interpreting them as discrete symbols rather than grounded spatial cues. This tokenization disrupts geometric continuity and visual context, leading to unreliable point reasoning and noisy prompts. As shown in Fig. \ref{fig:intro} , the zero-shot textual coordinate reasoning exhibits an overall accuracy below $60$\%. Given these limitations, a pivotal question naturally arises:

\begin{quote}
\textbf{(Q)} \textit{How can we select semantically informative point prompts and verify their correctness in a way that is grounded in visual--semantic alignment, rather than relying on coarse region priors or text-only coordinate reasoning?}
\end{quote}

To answer this question, we treat the coarse bounding box not merely as a spatial constraint, but as a \emph{structured uncertainty prior}. Regions within the box exhibit varying semantic relevance, suggesting that point prompts should be selected according to \emph{information gain} rather than spatial uniformity. Meanwhile, verifying a point's correctness should occur \emph{within the joint visual--language representation space}, rather than through coordinate text tokens—a design rationale strongly corroborated by our pilot study (Fig. \ref{fig:intro}), where vision-based reasoning significantly outperformed textual coordinate parsing in both accuracy and confidence under a zero-shot setting. Guided by these two principles, we propose a pairing of \emph{entropy-based discovery} and \emph{vision-based semantic validation}.

Guided by this insight, we propose \model, a framework for \emph{Referring Expression Segmentation via Entropy-based Prior Point Discovery with a Visual Reasoning Agent}. \model is built upon two key components: (1) an \textbf{Entropy-Based Point Discovery (EBD)} strategy that identifies high-information candidate points by modeling spatial uncertainty (Sec.~\ref{sec:entropy}), and (2) a \textbf{Vision-Based Reasoning (VBR)} module that verifies candidate correctness through visual--semantic alignment instead of textual coordinate inference (Sec.~\ref{sec:vision-reasoning}).

Building on these components, \model performs RES through a four-stage coarse-to-fine workflow: bounding box initialization (Sec.~\ref{sec:bbox}), entropy-guided point discovery (Sec.~\ref{sec:entropy}), vision-based point validation (Sec.~\ref{sec:vision-reasoning}), and final mask decoding with probability aggregation (Sec.~\ref{sec:decoder}). This formulation casts point prompting as an \emph{information maximization} process, improving grounding accuracy while requiring minimal prompts.

We evaluate \model on RefCOCO, RefCOCO+, RefCOCOg, and ReasonSeg. \model achieves new state-of-the-art performance across all benchmarks, demonstrating its ability to produce accurate and semantically grounded segmentations. Our contributions are summarized as follows:
\begin{itemize}
\item We propose an \textbf{Entropy-Based Point Discovery (EBD)} strategy that leverages spatial uncertainty to identify semantically critical prompt candidates.
\item We introduce a \textbf{Vision-Based Reasoning (VBR)} mechanism that verifies candidate correctness through visual--semantic alignment rather than text-only reasoning.
\item \model achieves state-of-the-art performance on RefCOCO, RefCOCO+, RefCOCOg, and ReasonSeg.
\end{itemize}

\section{Related Work}
\subsection{Referring Expression Segmentation}
Referring Expression Segmentation (RES)~\cite{hu2016Segmentation, kazemzadeh-etal-2014-referitgame, Mao_2016_CVPR, yu2016Modeling, Luo_2020_CVPR} evolves from its precursor, Referring Expression Comprehension (REC)~\cite{hu2016natural, liu2019Learning, Yang_2022_CVPR, wang2019Neighbourhood, yang2020Improving, yang2019fast}, by advancing from coarse bounding box localization to precise pixel-level segmentation. This shift places stricter demands on fine-grained visual-linguistic alignment. Early methods primarily depended on convolutional architectures~\cite{chen2019see, feng2021encoder, huang2020referring, liu2017recurrent} or vanilla attention mechanisms~\cite{ding2021vision, ding2022vlt, kim2022restr, liu2023caris, liu2023multi}, which often inadequately captured complex cross-modal interactions. Recent studies have enhanced segmentation performance by incorporating more sophisticated mechanisms into the models, such as the cross-attention in ReLA~\cite{Liu_2023_CVPR}, the vision-language token fusion in RefSegformer~\cite{wu2024toward}, and the counting-aware hierarchical decoding in CoHD~\cite{Luo_2025_cohd}. Nevertheless, these methods still struggle to address the intricate challenges posed by RES, especially failing to accurately capture the nuanced relationships between textual referring and their corresponding object regions.

\subsection{Pixel Understanding with MLLMs}

Pioneering the integration of Multimodal Large Language Models (MLLMs) with segmentation tasks, LISA~\cite{lai2024lisa} established the "one embedding as all mask" paradigm by augmenting the MLLM vocabulary with a special [SEG] token to elicit masks from decoders like SAM~\cite{kirillov2023segment,ravi2024sam}. Building upon this, subsequent works like GSVA~\cite{xia2024gsva} extend this ideal to address more complex scenarios, such as handling the rejection of null targets through a [REJ] token. Other efforts enhance versatility through dedicated decoding mechanisms, such as GLaMM's integration of masks with language responses for multi-granular prompts~\cite{Rasheed2024glamm} and PSALM's unified framework with an external decoder for diverse tasks~\cite{zhang2025PSALM}.

Unfortunately, most of these methods merely treat MLLMs as feature encoders, failing to leverage their robust reasoning capabilities. To address this, SAM4MLLM~\cite{Chen2025sam4mllm} seamlessly integrates MLLMs with SAM, leveraging the former’s dialog capability to generate effective prompt points for refining masks. SegAgent~\cite{zhu2025segAgent} further advances this direction by modeling segmentation as a multi-step Markov Decision Process, enabling MLLMs to iteratively generate text-based coordinates and imitate human annotators’ interactive reasoning. However, a common limitation of these approaches is their reliance on converting inherently visual reasoning tasks (e.g., point attribution) into textual form. This practice prevents MLLMs from utilizing their full reasoning potential and introduces noisy prompts that degrade segmentation accuracy. In contrast, our \model incorporates explicit visual reasoning, fully utilizing MLLMs’ VQA capabilities to mitigate textualized coordinate limitations.

\section{Method}
\label{sec:method}

\begin{figure*}[t]
  \centering
   \includegraphics[width=\linewidth]{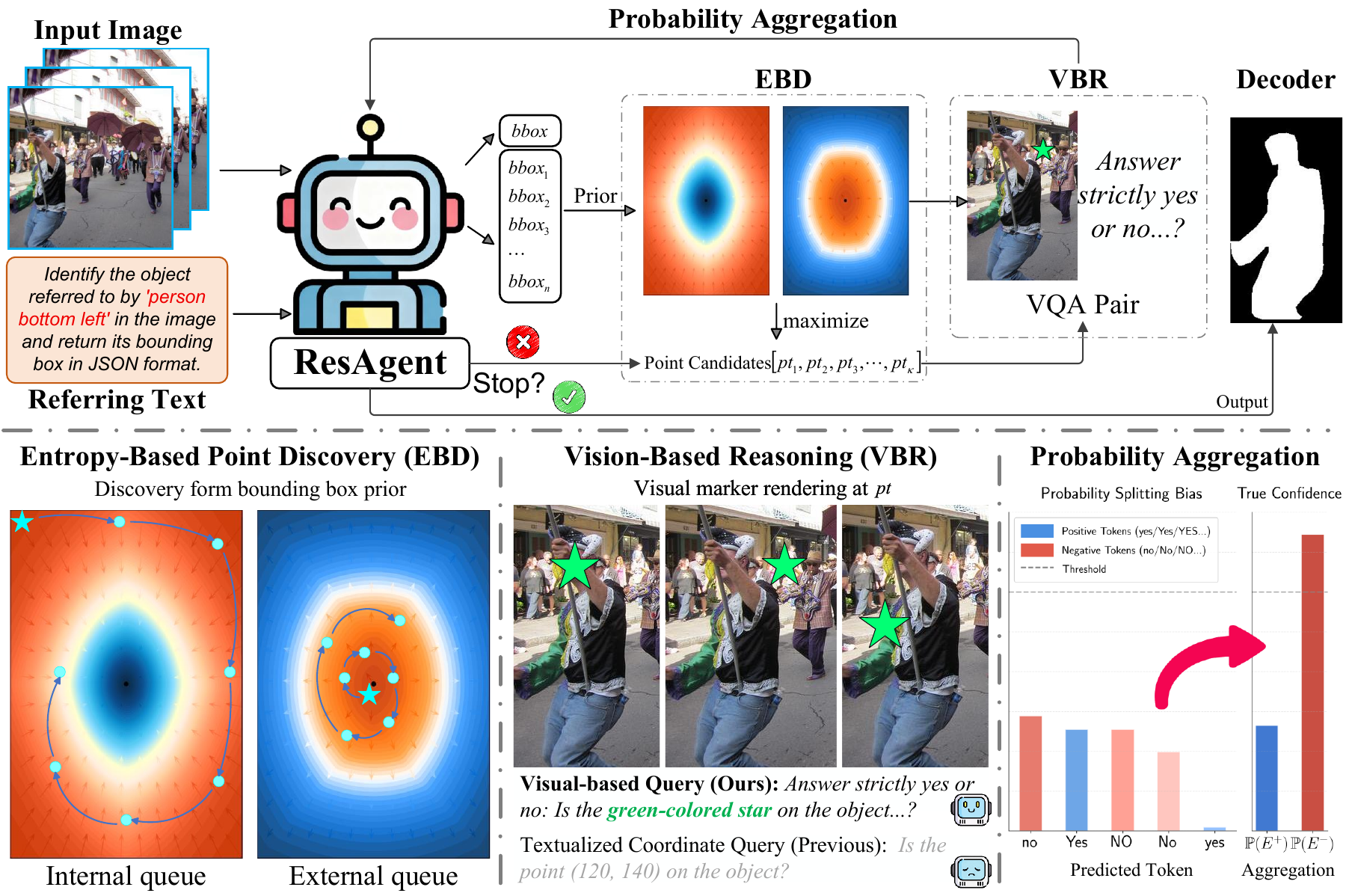}

   \caption{Overall framework of \model. Starting with an MLLM-inferred bounding box as prior, the framework first uses Entropy-Based Point Discovery (EBD) to select high-value candidate points. Then \model employs Vision-Based Reasoning (VBR) to validate points through visual markers and VQA queries, filtering noise. Early stopping is triggered when sufficient qualified points are collected, which are fed into the SAM decoder to generate accurate segmentation masks.}
   \label{fig:framework}
   \vspace{-3mm}
\end{figure*}



Referring expression segmentation (RES) requires aligning textual descriptions with precise pixel regions. Existing MLLM-based approaches~\cite{Chen2025sam4mllm, zhu2025segAgent} predominantly adopt random or uniform strategies for point prompt selection and encode point coordinates as textual tokens. These strategies face two major challenges. First, sampling a large number of points (e.g., $25$ in SAM4MLLM~\cite{Chen2025sam4mllm}) increases inference complexity and latency while introducing redundancy and noise into the prompts. Second, numeric coordinates are split into character-level tokens (e.g., $175$ $\rightarrow$ `1', `7', `5'), fragmenting spatial semantics, obscuring absolute and relative positional information, and often leading to inconsistent positive/negative point selection or target misidentification. Together, these issues degrade an MLLM’s ability to reason over structured spatial relations and constrain the overall performance of RES.



To overcome these limitations (Fig.~\ref{fig:framework}), we propose \textbf{\model}---\emph{Referring Expression Segmentation via Entropy-based Prior Point Discovery with a Visual Reasoning Agent}. \model integrates an MLLM with a SAM-based segmentation module and executes a four-stage, coarse-to-fine collaborative workflow (overviewed in Secs.~\ref{sec:task}--\ref{sec:decoder}):

Sec.~\ref{sec:bbox}: \textbf{Bounding Box Prediction} --- an MLLM predicts a coarse localization to initialize the search region.
 Sec.~\ref{sec:entropy}: \textbf{Entropy-Based Point Discovery} --- high-value candidates are identified via an uncertainty field inside the coarse box.
 Sec.~\ref{sec:vision-reasoning}: \textbf{Vision-Based Point Validation} --- the MLLM’s multimodal reasoning verifies candidate correctness in the visual domain.
 Sec.~\ref{sec:decoder}: \textbf{Mask Decoding and Model Adaptation} --- refined segmentation is generated and modules are adapted for robustness.
This design reformulates RES as a coarse-to-fine, information-theoretic reasoning process that maximizes uncertainty reduction and aligns visual and linguistic representations with minimal prompt overhead.

\subsection{Task Definition}
\label{sec:task}

RES seeks pixel-level segmentation guided by text semantics and involves three key components: (1) an input RGB image $\mathcal{I}$ containing target objects; (2) a textual expression $\mathcal{T}$ describing target attributes (e.g., category, shape, relations); and (3) a referential mask $\mathcal{M}$ semantically consistent with $\mathcal{T}$, where foreground pixels denote the referred region. Formally, given joint input $(\mathcal{I},\mathcal{T})$, the model outputs a mask $\mathcal{M}$:
\begin{equation}
\mathcal{M} \;=\; \Theta(\mathcal{I}, \mathcal{T}),
\end{equation}
where $\Theta$ denotes the RES model.

\subsection{Bounding Box Prediction}
\label{sec:bbox}

As the first stage of \model (Fig.~\ref{fig:framework}), the bounding box prediction module obtains a coarse spatial prior corresponding to $\mathcal{T}$, initializing the search region for entropy-based point discovery (Sec.~\ref{sec:entropy}). We design a task-specific instruction $\mathrm{INSTR}$ to elicit a structured spatial response:
\begin{quote}
\emph{``Identify the object referred to by `\{$\mathcal{T}$\}' in the image and return its bounding box in JSON format.''}
\end{quote}
Feeding $(\mathcal{I}, \mathrm{INSTR}(\mathcal{T}))$ into the MLLM yields a predicted box,
\begin{equation}
    \mathrm{bbox} \;=\; \mathrm{MLLM}\big(\mathcal{I}, \mathrm{INSTR}(\mathcal{T})\big).
\end{equation}
MLLMs typically return \emph{relative} coordinates. For any $(x,y)$ in an image of width $w$ and height $h$, the model may encode positions as $\big(\tfrac{x}{w}\!\times\!\alpha, \tfrac{y}{h}\!\times\!\alpha\big)$ with a scaling factor $\alpha$ (e.g., $\alpha=1000$). We recover \emph{absolute} pixels via an inverse mapping:
\begin{equation}
    \mathrm{bbox}^{\prime} \;=\; \mathrm{convert}^{-1}(\mathrm{bbox}).
\end{equation}

\subsection{Entropy-Based Point Discovery}
\label{sec:entropy}

This core stage refines $\mathrm{bbox}^{\prime}$ by selecting candidate points that maximize \emph{information gain} for target membership ($E^+$: object, $E^-$: background). The central idea is to cast point discovery as uncertainty maximization under a spatial prior.

\paragraph{Structured Uncertainty Prior.}
Empirically, the spatial distribution inside $\mathrm{bbox}^{\prime}$ is not uniform: central regions are more likely object-affiliated (potential positives), while corners/outer regions are more likely background (potential negatives). Let $d_c(\mathrm{pt})$ be the distance from a point to the box center and $d_e(\mathrm{pt})$ the distance to the nearest edge; these distances induce a \emph{structured} uncertainty field.

\paragraph{Bernoulli Entropy Formulation.}
We model target membership at $\mathrm{pt}$ as a Bernoulli variable with probability $p(\mathrm{pt}) = \mathbb{P}(E^+|\mathrm{pt})$. Rather than using raw distances as entropy surrogates, we \emph{calibrate} $p(\mathrm{pt})$ from geometry via a logistic mapping
\begin{equation}
\label{eq:entropy_eq}
p(\mathrm{pt}) \;=\; \sigma\!\Big(a - b\,\tilde{d}_c(\mathrm{pt}) + c\,\tilde{d}_e(\mathrm{pt})\Big),
\quad \text{with} \quad
\tilde{d}_c, \tilde{d}_e \in [0,1],
\end{equation}
where $\sigma$ is the sigmoid function and $(a,b,c)$ are hyperparameters; $\tilde{d}_c,\tilde{d}_e$ are distances normalized within $\mathrm{bbox}^{\prime}$. Following Shannon's information theory~\cite{Shannon1948entropy}, higher unpredictability yields greater information content. The \emph{Shannon entropy} at $\mathrm{pt}$ is
\begin{equation}
\label{eq:h_entropy}
H(\mathrm{pt}) \;=\; -\,p(\mathrm{pt})\log p(\mathrm{pt}) - \big(1-p(\mathrm{pt})\big)\log\!\big(1-p(\mathrm{pt})\big),
\end{equation}
which is maximized near $p(\mathrm{pt})\!\approx\!0.5$. This principled form subsumes the intuitive rule that \emph{boundary-proximal positives} and \emph{center-proximal negatives} are the most informative.

\paragraph{Practical entropy proxy and sampling.}
Direct computation of $H(\mathrm{pt})$ from the MLLM is infeasible; we therefore instantiate the calibrated entropy field in image space via a geometry-driven proxy and a compact sampling operator. Concretely, a parameterized superellipse spiral is generated inside $\mathrm{bbox}^{\prime}$ and adaptive arc-length sampling is performed, with local density modulated by the numerical gradient of the normalized radial distance (an analytically justified surrogate for $\nabla H$ in Eq.~\ref{eq:h_entropy}). Sampled locations are scored by $p(\mathrm{pt})$ and ranked by $H(\mathrm{pt})$ to yield the candidate pool for visual verification. Full details are given in appendix.

\paragraph{Dual-Queue Greedy Discovery.}
We realize efficient high-entropy search with two queues:
\begin{itemize}
    \item \textbf{Internal queue}: scans inward from the four edges (top/bottom/left/right) along inward normals to harvest boundary-proximal candidates.
    \item \textbf{External queue}: expands outward from the center along horizontal/vertical rays to harvest center-proximal candidates.
\end{itemize}
We rank visited locations by $H(\mathrm{pt})$ and retain the top-$\mathcal{K}$ for verification (Sec.~\ref{sec:vision-reasoning}), where $\mathcal{K}$ is the candidate budget. This procedure is equivalent in spirit to the distance-proportional heuristic, but expressed in a calibrated entropy space (consistent with the information-theoretic motivation).

\subsection{Vision-Based Reasoning for Point Inference}
\label{sec:vision-reasoning}

We verify high-entropy candidates using \emph{direct visual reasoning} rather than textualized coordinates. The MLLM’s VQA capability is used to infer membership ($E^+$/$E^-$) in the visual domain, avoiding semantic loss from tokenized numbers.

\paragraph{Reasoning Protocol.}
For each candidate $\mathrm{pt}$:
\textbf{Visual marker rendering:} overlay a predefined marker (e.g., star, circle) at $\mathrm{pt}$ on $\mathcal{I}$. Assign distinct colors from a diverse palette and record each color name to prevent ambiguity.
\textbf{Structured query:} prompt the MLLM with a constrained binary question:
    \begin{quote}
    \emph{``Answer strictly yes or no: Is the \{\textit{color}\}-colored \{\textit{marker}\} on the object referred to by `\{$\mathcal{T}$\}' in the picture?''}
    \end{quote}
\paragraph{Probability Aggregation for Robustness.}
To mitigate \emph{probability splitting} across semantically identical surface forms (``yes''/``Yes'', ``no''/``No''), we aggregate pre-decoding token probabilities:
\begin{equation}
\mathbb{P}(E^+) \;=\; \sum_{\text{yes-variants}} p_t,
\qquad
\mathbb{P}(E^-) \;=\; \sum_{\text{no-variants}} p_t,
\end{equation}
using the Top-$k$ logits (we adopt $k=5$). Points with confidence above $\eta$ are retained. This substantially improves decision stability compared to textualized coordinate reasoning or naive single-token decoding.

\paragraph{Early Stopping.}
Inference halts once a sufficient set of high-confidence points is collected---two positives ($E^+$) and one negative ($E^-$)---eliminating redundant queries while preserving accuracy.

\subsection{Decoder and Adaptation}
\label{sec:decoder}

\paragraph{Mask Decoding.}
The SAM module receives $(\mathcal{I}, \mathrm{bbox}^{\prime})$, two high-confidence positives $\mathbf{pt}^{+}$, and one high-confidence negative $\mathbf{pt}^{-}$:
\begin{equation}
\mathcal{M}^{\prime} \;=\; \mathrm{SAM}\big(\mathcal{I},\, \mathrm{bbox}^{\prime},\, \mathbf{pt}^{+},\, \mathbf{pt}^{-}\big),
\end{equation}
producing a mask $\mathcal{M}^{\prime}$ aligned with $\mathcal{T}$.

\paragraph{Model Adaptation.}
We strengthen the MLLM’s visual point inference via LoRA-based fine-tuning. Training samples are constructed with $2\times$ the inference point count; marker colors and shapes are randomized to improve generalization. During inference, randomly colored stars are used for robustness.

Following Chen et al.~\cite{Chen2025sam4mllm}, we observe an annotation-style discrepancy between COCO and SAM pretraining (SA-1B). Therefore, SAM is separately fine-tuned to the COCO annotation style, mitigating segmentation bias due to label inconsistency.

\section{Experiments}

\subsection{Dataset}
We employ four challenging referring expression segmentation benchmarks in our experiments:

\textbf{RefCOCO(/+/g)}\cite{kazemzadeh-etal-2014-referitgame,Mao_2016_CVPR} are widely used benchmarks for referring segmentation, all built upon images from the COCO\cite{lin2014microsoft} dataset. RefCOCO features relatively simple referring expressions, while refCOCO+ prohibits location-based descriptions to focus more on object appearance and attributes. RefCOCOg further includes longer and more detailed expressions, increasing task complexity. Following previous works, evaluation adopts mean Intersection-over-Union (mIoU, average IoU across samples) as core metric.

\textbf{ReasonSeg}\cite{lai2024lisa} is a dedicated benchmark for \textbf{reasoning segmentation}, which requires models to generate segmentation masks from implicit, reasoning-intensive text instructions rather than explicit object descriptions. Its images are sourced from OpenImages\cite{Kuznetsova2020openImage} and ScanNetv2\cite{dai2017scanNet}, with image-instruction pairs (accompanied by high-quality masks) designed for complexity. Instructions include short phrases (e.g., "non-spicy food") and long sentences, all demanding world knowledge or multi-step reasoning. Following the original design, evaluation adopts gIoU and cIoU as core metrics—gIoU is calculated as the average of all per-image IoUs, while cIoU is defined as the cumulative intersection over the cumulative union.


\subsection{Main Results}

\begin{table*}[htbp]
\small
\centering
\setlength{\tabcolsep}{3pt}
\caption{Comparison with state-of-the-art methods on image referring expression segmentation (RES) and reasoning segmentation datasets, including RefCOCO/+/g \cite{kazemzadeh-etal-2014-referitgame,Mao_2016_CVPR} and ReasonSeg \cite{lai2024lisa} (\texttt{val}). The best and second-best results are marked \textbf{bold} and \underline{underlined}, respectively.}
\begin{tabularx}{0.95\linewidth}{
  l  
  |c  
  |*{3}{>{\centering\arraybackslash}X}  
  |*{3}{>{\centering\arraybackslash}X}  
  |*{2}{>{\centering\arraybackslash}X}  
  |*{2}{>{\centering\arraybackslash}X}  
}
\toprule
\multirow{2}{*}{\textbf{Method}} & \multirow{2}{*}{\textbf{Size}} & \multicolumn{3}{c|}{\textbf{RefCOCO}} & \multicolumn{3}{c|}{\textbf{RefCOCO+}} & \multicolumn{2}{c|}{\textbf{RefCOCOg}} & \multicolumn{2}{c}{\textbf{ReasonSeg}} \\
\cmidrule(lr){3-5} \cmidrule(lr){6-8} \cmidrule(lr){9-10} \cmidrule(lr){11-12}
&& val & testA & testB & val & testA & testB & val(U) & test(U) & gIoU & cIoU \\
\midrule
\rowcolor{gray!10}\multicolumn{12}{l}{\textit{\textcolor{gray}{Non-LLM-based Specialists}}} \\
\text{ReLA}$_{\text{[CVPR'23]}}$ \cite{Liu_2023_CVPR} & -- & 73.8 & 76.5 & 70.2 & 66.0 & 71.0 & 57.7 & 65.0 & 66.0 & 22.4 & 19.9\\
 \text{X-Decoder}$_{\text{[CVPR'23]}}$ \cite{zou2023xdecoder} & -- & -- & -- & -- & -- & -- & -- & 64.6 & -- & 22.6 & 17.9 \\
 \text{SEEM}$_{\text{[NIPS'23]}}$ \cite{zou2023seem} & -- & -- & -- & -- & -- & -- & -- & 65.7 & -- & 25.5 & 21.2 \\
  \text{CoHD}$_{\text{[ICCV'25]}}$ \cite{Luo_2025_cohd} &-- & 78.11 &80.39 &75.20 &72.03& 76.37 &65.45 &70.83 &72.11  & -- & --\\
\midrule

\rowcolor{gray!10}\multicolumn{12}{l}{\textit{\textcolor{gray}{LLM-based Image Generalists}}} \\
 \text{NExT-Chat}$_{\text{[ICML'24]}}$ \cite{zhang2024nextChat} & 7B & 74.7 & 78.9 & 69.5 & 65.1 & 71.9 & 56.7 & 67.0 & 67.0 & -- & -- \\
 \text{PixelLM}$_{\text{[CVPR'24]}}$ \cite{Ren_2024_pixellm} & 7B & 73.0 & 76.5 & 68.2 & 66.3 & 71.7 & 58.3 & 69.3 & 70.5 & -- & -- \\
 \text{LISA}$_{\text{[CVPR'24]}}$ \cite{lai2024lisa} & 7B & 74.9 & 79.1 & 72.3 & 65.1 & 70.8 & 58.1 & 67.9 & 70.6 & 61.3 & \underline{62.9} \\
 \text{Groundhog}$_{\text{[CVPR'24]}}$ \cite{Zhang_2024_groundHOG} & 7B & 78.5 & 79.9 & 75.7 & 70.5 & 75.0 & 64.9 & 74.1 & 74.6 & 56.2 & -- \\
  \text{LaSagnA}$_{\text{[arXiv'24]}}$\cite{wei2024lasagna} & 7B & 76.8 & 78.7 & 73.8 & 66.4 & 70.6 & 60.1 & 70.6 & 71.9 & 48.8 & 47.2 \\

\text{GSVA}$_{\text{[CVPR'24]}}$\cite{xia2024gsva} & 8B & 77.2 &78.9 &73.5& 65.9& 69.6& 59.8& 72.7& 73.3 & -- & --\\

\text{GSVA}$_{\text{[CVPR'24]}}$\cite{xia2024gsva} & 13B & 79.2 &81.7 &77.1 &70.3 &73.8 &63.6 &75.7 &77.0 & -- & --\\
  
\text{SAM4MLLM}$_{\text{[ECCV'24]}}$\cite{Chen2025sam4mllm} & 8B & 79.8 & 82.7 & 74.7 & 74.6 & 80.0 & 67.2 & 75.5 & 76.4 & 58.4 & 60.4 \\
  \text{M$^{\rm 2}$SA}$_{\text{[ICLR'25]}}$ \cite{jang2025mmr} & 13B & 74.6 & 77.6 & 71.0 & 64.0 & 68.1 & 57.6 & 69.0 & 69.3 & -- & -- \\
 \text{SegAgent}$_{\text{[CVPR'25]}}$\cite{zhu2025segAgent} & 7B & 79.69 & 81.35 & 76.57 & 72.49 & 75.80 & 66.89 & 75.11 & 75.20 & -- & -- \\
 \text{Text4Seg}$_{\text{[ICLR'25]}}$\cite{lan2025textseg} & 8B & 79.2 & 81.7 & 75.6 & 72.8 & 77.9 & 66.5 & 74.0 & 75.3 & -- & -- \\
 \text{Text4Seg}$_{\text{[ICLR'25]}}$\cite{lan2025textseg} & 13B & 80.2 & \underline{82.7} & 77.3 & 73.7 & 78.6 & 67.6 & 74.0 & 75.1 & -- & -- \\
\midrule
\rowcolor{gray!10}\multicolumn{12}{l}{\textit{\textcolor{gray}{LLM-based Video Generalists}}} \\
 \text{VideoLISA}$_{\text{[NIPS'24]}}$ \cite{bai2024videoLisa} & 3.8B & 73.8 & 76.6 & 68.8 & 63.4 & 68.8 & 56.2 & 68.3 & 68.8 & 61.4 & \underline{67.1} \\
 \text{VISA}$_{\text{[ECCV'24]}}$ \cite{yan2024visa} & 7B & 72.4 & 75.5 & 68.1 & 59.8 & 64.8 & 53.1 & 65.5 & 66.4 & 52.7 & 57.8 \\
 \text{Vitron}$_{\text{[NIPS'24]}}$ \cite{fei2024vitron} & 7B & 75.5 & 79.5 & 72.2 & 66.7 & 72.5 & 58.0 & 67.9 & 68.9 & -- & -- \\
\text{Sa2VA}$_{\text{[arXiv'25]}}$ \cite{yuan2025sa2va} & 4B & 78.9 & -- & -- & 71.7 & -- & -- & 74.1 & -- & -- & -- \\
\midrule

\rowcolor{blue!7.5} \textbf{\model} (Ours) & 4B & \underline{80.50} & \bf 82.75 & \bf 78.32 & \underline{75.13} & \underline{80.10} & \underline{69.23} & \underline{76.89} & \underline{77.97} & \bf 72.74 & \bf 68.38 \\
\rowcolor{blue!7.5} \textbf{\model} (Ours) & 8B & \textbf{81.19} & 82.64 & \underline{78.07} & \textbf{75.88} & \textbf{80.24} & \textbf{70.01} & \textbf{77.23} & \textbf{78.04} & \underline{70.51} & 66.86 \\
\bottomrule
\end{tabularx}
\label{tab:main_results}
\vspace{-3mm}
\end{table*}

To validate the effectiveness of our \model for referring expression segmentation (RES), we conduct experiments on several popular benchmark datasets: RefCOCO/+/g\cite{kazemzadeh-etal-2014-referitgame,Mao_2016_CVPR} (\texttt{val}, \texttt{testA}, \texttt{testB} subsets) and ReasonSeg\cite{lai2024lisa} (\texttt{val}). We compare our \model with various state-of-the-art (SOTA) approaches, which are categorized into three groups: \textit{Non-LLM-based Specialists}, \textit{LLM-based Image Generalists}, and \textit{LLM-based Video Generalists}. As presented in Table \ref{tab:main_results}, our \model achieves remarkable performance across all benchmark datasets, outperforming state-of-the-art methods in three distinct categories.

For \textit{Non-LLM-based Specialists}, CoHD\cite{Luo_2025_cohd} represents the top performer on RefCOCO (\texttt{val}, \texttt{testA}, \texttt{testB} subsets). Our $4$B \model surpasses it by $2.39$\%, $2.36$\%, $3.12$\% mIoU on these benchmarks, while our $8$B version further extends the lead to $3.08$\%, $2.25$\%, $2.87$\% mIoU. This result demonstrates that integrating LLM-driven reasoning into RES tasks yields substantial gains over specialized non-LLM architectures.

Among \textit{LLM-based Image Generalists}—the most competitive category—SAM4MLLM\cite{Chen2025sam4mllm} ($8$B) and Text4Seg\cite{lan2025textseg} ($13$B) stand out, with the latter achieving $82.7$\% mIoU on RefCOCO \texttt{testA}. Our $4$B \model not only matches this performance ($82.75$\% mIoU) but also sets a new record of $78.32$\% mIoU on RefCOCO \texttt{testB} (surpassing Text4Seg’s $77.3$\% by $1.02$\% mIoU). Notably, our $8$B \model outperforms all counterparts across RefCOCO+ ($75.88$\%, $80.24$\%, $70.01$\% mIoU on \texttt{val}, \texttt{testA}, \texttt{testB}) and RefCOCOg ($77.23$\%, $78.04$\% mIoU on \texttt{val} and \texttt{testU}), despite having a smaller parameter size than $13$B competitors like GSVA\cite{xia2024gsva} and Text4Seg—underscoring the effectiveness of our point discovery method combined with visual reasoning for RES tasks.

In the \textit{LLM-based Video Generalists} category, VideoLISA\cite{bai2024videoLisa} ($3.8$B) delivers the best results on ReasonSeg with $67.1$\% cIoU. Our $4$B \model outperforms it by $1.28$\% ($68.38$\% cIoU) and achieves a striking $72.74$\% gIoU on ReasonSeg—surpassing all methods in this category by over $11$\%. This highlights the effectiveness of our visual reasoning strategy in handling complex semantic inference, even compared to models optimized for video-level temporal reasoning.

Across all datasets, our \model consistently excels in challenging subsets (e.g., RefCOCO \texttt{testB} with cluttered backgrounds and RefCOCOg (\texttt{val}/\texttt{test}) with complex textual descriptions), validating the rationality of our information entropy-based point discovery and visual marker-guided reasoning design.

\subsection{Ablation Study}
\begin{table}[htbp]
    \centering
    \small
    \begin{tabular}{l|ccc|c}
        \toprule
        \multirow{2}{*}{\bf Method} & \multicolumn{3}{c|}{\bf RefCOCO} & \multirow{2}{*}{\bf Avg.} \\
        \cmidrule(lr){2-4}
        & \bf Val & \bf TestA & \bf TestB &  \\
        \midrule
        Baseline                  & 77.09 & 79.79 & 72.89 & 76.59 \\
        + EBD                     & 78.74 & 80.50 & 73.42 & 77.55 \\
        + EBD + VBR               & 79.19 & 81.61 & 75.24 & 78.68 \\
        + EBD + VBR + PA          & 79.76 & 82.16 & 75.58 & 79.17 \\
        \textbf{\model (Full) }            & \bf 81.19 & \bf 82.64 & \bf 78.07 & \bf 80.63 \\
        \bottomrule
    \end{tabular}
    \caption{Ablation study on core components of ResAgent.}
    \vspace{-2mm}
    \label{tab:ablation_core}
\end{table}

To verify the contributions of each core component in our \model, we conduct systematic ablation experiments on the RefCOCO dataset, with the component setup and results detailed in Table \ref{tab:ablation_core}. Our ablation study starts with a straightforward baseline: we feed the bounding box ($bbox^{\prime}$) inferred by an MLLM into SAM for segmentation without any additional optimization. We then incorporate three key components—Information \textbf{E}ntropy-\textbf{B}ased Point \textbf{D}iscovery (\textbf{EBD}), \textbf{V}ision-\textbf{B}ased \textbf{R}easoning (\textbf{VBR}), and \textbf{P}robability \textbf{A}ggregation (\textbf{PA})—to evaluate their individual and synergistic effects sequentially.

Starting with a baseline average mIoU (Avg.) of $76.59$\% on RefCOCO, incorporating EBD with traditional textualized coordinate inference first boosts the Avg. to $77.55$\% (a $0.96$\% mIoU improvement), demonstrating that EBD effectively identifies regions with rich semantic information, reduces redundant points, and provides SAM with more targeted guidance compared to random or uniform sampling. Further integrating VBR elevates the Avg. to $78.68$\% (an additional $1.13$\% mIoU gain), validating that vision-based reasoning filters out noisy points and mitigates mask inaccuracies caused by misleading inputs. Adding PA then pushes the Avg. to $79.17\%$ (a $0.49$\% mIoU increase), as probability aggregation consolidates the confidence of validated points, minimizing the uncertainty of individual point judgments and stabilizing segmentation performance. Finally, our full \model—combining all three components with additional fine-tuning adaptations—achieves a standout Avg. of $80.63\%$ ($4.04\%$ mIoU higher than the baseline) and outperforms all ablated variants across the three RefCOCO subsets.

\subsection{Analysis}

\begin{table}[htbp]
    \centering
    \setlength{\tabcolsep}{3pt}
    \small
    \begin{tabular}{lcccc}
        \toprule
        \multirow{2}{*}{\bf Reasoning Strategy} & \multicolumn{3}{c}{\bf RefCOCO} & \multirow{2}{*}{\bf Avg.} \\
        \cmidrule(lr){2-4}
        & \bf Val & \bf TestA & \bf TestB & \\
        \midrule
        Textual Coordinate & 79.04 &81.32 &74.26 &78.20 \\
        \rowcolor{blue!7.5} \bf VBR (Ours) & \bf 81.19 & \bf 82.64 & \bf 78.07& \bf 80.63 \\
        \bottomrule
    \end{tabular}
    
    \caption{Comparison of different point reasoning strategies of full model.}
    \vspace{-3mm}
    \label{tab:reasoning_strategy}
\end{table}

\textbf{Impact of different point reasoning strategies.} To isolate the effect of point reasoning mechanisms (Sec. \ref{sec:vision-reasoning}), we conduct experiments on the full \model by exclusively replacing the reasoning strategy—comparing the traditional \textit{Textual Coordinate} approach with our proposed \textit{Vision-Based Reasoning} (VBR). The former converts point coordinates into textual sequences for reasoning, relying solely on isolated textual descriptions (losing spatial context) to validate point reliability. As shown in Table \ref{tab:reasoning_strategy}, our VBR outperforms it by $2.43\%$ average mIoU ($80.63\%$ vs. $78.20\%$), with the most notable gain of $3.81\%$ on RefCOCO \texttt{TestB}. This improvement stems from VBR’s ability to align points with the target’s visual-semantic features via markers and VQA—effectively handling cluttered or ambiguous RES scenarios.


\begin{table}[t]
    \centering
    \small
    \begin{tabular}{lccccc}
        \toprule
        \multirow{2}{*}{\bf $\mathcal{K}$ } & \multicolumn{3}{c}{\bf RefCOCO} & \multirow{2}{*}{\bf Avg. ($\uparrow $)} & \multirow{2}{*}{\bf Time ($\downarrow $)} \\
        \cmidrule(lr){2-4}
        & \bf Val & \bf TestA & \bf TestB & & \\
        \midrule
        \rowcolor{gray!10} \textcolor{gray}{0} & \textcolor{gray}{77.09} & \textcolor{gray}{79.79} & \textcolor{gray}{72.89} & \textcolor{gray}{76.59} & \textcolor{blue}{4.03 s} \\
        2 & 79.16 & 81.56 & 74.69 & 78.47 & 4.27 s\\
        3 & 79.45 & 82.04 & 75.58 & 79.02 & 4.54 s \\
        4 & \bf 81.19 & \bf 82.64 & \bf 78.07 & \bf 80.63 & 4.86 s \\
        5 & 80.17 & 81.56 & 77.70 & 79.87 & 5.03 s\\
        \bottomrule
    \end{tabular}
    
    \caption{Parameter analysis of candidate point number $\mathcal{K}$.}
    \vspace{-1mm}
    \label{tab:param_k}
\end{table}

\textbf{Impact of candidate point number.} We analyze the influence of candidate point number $\mathcal{K}$ on performance and inference efficiency (Sec. \ref{sec:entropy}), with results presented in Table \ref{tab:param_k} ($\mathcal{K}=0$ denotes the baseline without additional candidate points). As $\mathcal{K}$ increases from $2$ to $4$, the average mIoU rises steadily—from $78.47\%$ to $\bf 80.63\%$—with consistent gains across all RefCOCO subsets, as more candidate points enhance coverage of the target’s semantic regions. However, when $\mathcal{K}$ further increases to $5$, the average mIoU drops to $79.87\%$ while inference time climbs to $5.03$ s ($19.3$\% slower than $\mathcal{K}=0$), as excessive points introduce redundant noise that misleads SAM and increases computational overhead. Notably, $\mathcal{K}=4$ achieves the optimal balance between accuracy and efficiency: it delivers the highest performance (e.g., $82.64\%$ mIoU on \texttt{testA} and $78.07\%$ on \texttt{testB}) with a moderate time cost of $4.86$ s, validating that our EBD-driven point inference works best with a sufficient yet non-redundant number of candidates.

\begin{table}[t]
    \centering
    \small
    \begin{tabular}{lcccc}
        \toprule
        \multirow{2}{*}{\bf (pos, neg)} & \multicolumn{3}{c}{\bf RefCOCO} & \multirow{2}{*}{\bf Avg.} \\
        \cmidrule(lr){2-4}
        & \bf Val & \bf TestA & \bf TestB & \\
        \midrule
        (1,1) & 79.16 & 81.56 & 74.69 & 78.47 \\
        (1,2) & 78.84 & 81.32 & 74.40 & 78.19 \\
        \rowcolor{blue!7.5} (2,1) & \bf 79.45 & \bf 81.83 & \bf 75.35 & \bf78.88 \\
        (2,2) & 79.04 & 81.50 & 74.69 & 78.41 \\
        (3,2) & 79.26 & 81.60 & 75.34 & 78.73 \\
        (2,3) & 78.35 & 81.00 & 74.21 & 77.85 \\
        \bottomrule
    \end{tabular}
    
    \caption{Parameter analysis of (pos, neg) combinations.}
    \vspace{-4mm}
    \label{tab:param_pos_neg}
\end{table}

\textbf{Impact of positive/negative combination.} In Sec. \ref{sec:vision-reasoning}, the $2$ positive ($E^+$) with $1$ negative ($E^-$) point combination served as the criterion for Early Stopping. To systematically investigate the impact of this hyperparameter, we evaluate segmentation accuracy across different (pos, neg) point combinations, as reported in Table \ref{tab:param_pos_neg}. Among all tested combinations, $(2,1)$ achieves the optimal average mIoU of $78.88\%$, outperforming other setups across RefCOCO all subsets. This superiority stems from a balanced trade-off between target coverage and background suppression: $2$ positive points sufficiently capture key semantic regions of the target, while $1$ negative point effectively filters out background interference without introducing redundant noise. In contrast, combinations with fewer positive points (e.g., $(1,1)$, $(1,2)$) fail to cover the target comprehensively, leading to incomplete segmentation. Those with excessive negative points (e.g., $(2,2)$, $(2,3)$) or redundant positive points (e.g., $(3,2)$) confuse SAM’s mask prediction by blurring the target-background boundary. These results validate that the $(2,1)$ positive-negative point combination is most compatible with our EBD and VBR pipeline, maximizing the reliability of point-guided segmentation.

\begin{table}[t]
    \centering
    \setlength{\tabcolsep}{3pt} 
    \small
    \begin{tabular}{lccccc}
        \toprule
        \multirow{2}{*}{\bf SAM Decoder} & \multirow{2}{*}{\bf Adapted?} & \multicolumn{3}{c}{\bf RefCOCO } & \multirow{2}{*}{\bf Avg.} \\
        \cmidrule(lr){3-5}
        & & \bf Val & \bf TestA & \bf TestB & \\
        \midrule
        \multirow{2}{*}{Base-Plus}  & \xmark  & 79.45 & 81.83 & 75.35 & 78.88 \\
                                & \cellcolor{blue!7.5}\cmark & \cellcolor{blue!7.5} 80.93 & \cellcolor{blue!7.5} 82.51 &  \cellcolor{blue!7.5} \bf78.08 & \cellcolor{blue!7.5} 80.51 \\
        \midrule
        \multirow{2}{*}{Large} & \xmark  & 79.76 & 82.16 & 75.58 & 79.17 \\
                               &\cellcolor{blue!7.5} \cmark & \cellcolor{blue!7.5}  \bf81.19 &  \cellcolor{blue!7.5} \bf82.64 &  \cellcolor{blue!7.5} 78.07 & \cellcolor{blue!7.5} \bf80.63 \\
        \bottomrule
    \end{tabular}
    \caption{Impact of style adaptation on SAM decoders (Base/Large). "Adapted?" denotes whether SAM is fine-tuned to fit COCO annotation style.}
    \label{tab:decoder_adaptation}
    \vspace{-3mm}
\end{table}

\textbf{Impact of decoder adaptation.} We analyze the effect of SAM decoder style adaptation (fine-tuning to fit COCO annotation style in Sec. \ref{sec:decoder}) across two decoder variants (`Base-Plus' and `Large'), with results shown in Table \ref{tab:decoder_adaptation}. For both decoders, adaptation yields consistent performance gains: the `Base-Plus' variant’s average mIoU rises from $78.88\%$ to $80.51\%$, while the `Large' variant improves from $79.17\%$ to $\bf 80.63\%$. This improvement stems from adaptation mitigating the distribution shift between SAM’s pre-trained weights and the target dataset’s annotation style—aligning the model’s mask generation with the dataset’s labeling conventions to reduce systematic inaccuracies. Notably, the `Large' decoder benefits more from adaptation, highlighting the synergistic effect between a more expressive decoder architecture and style alignment, which further refines fine-grained segmentation details.

\begin{figure}[htbp]
  \centering
   \includegraphics[width=\linewidth]{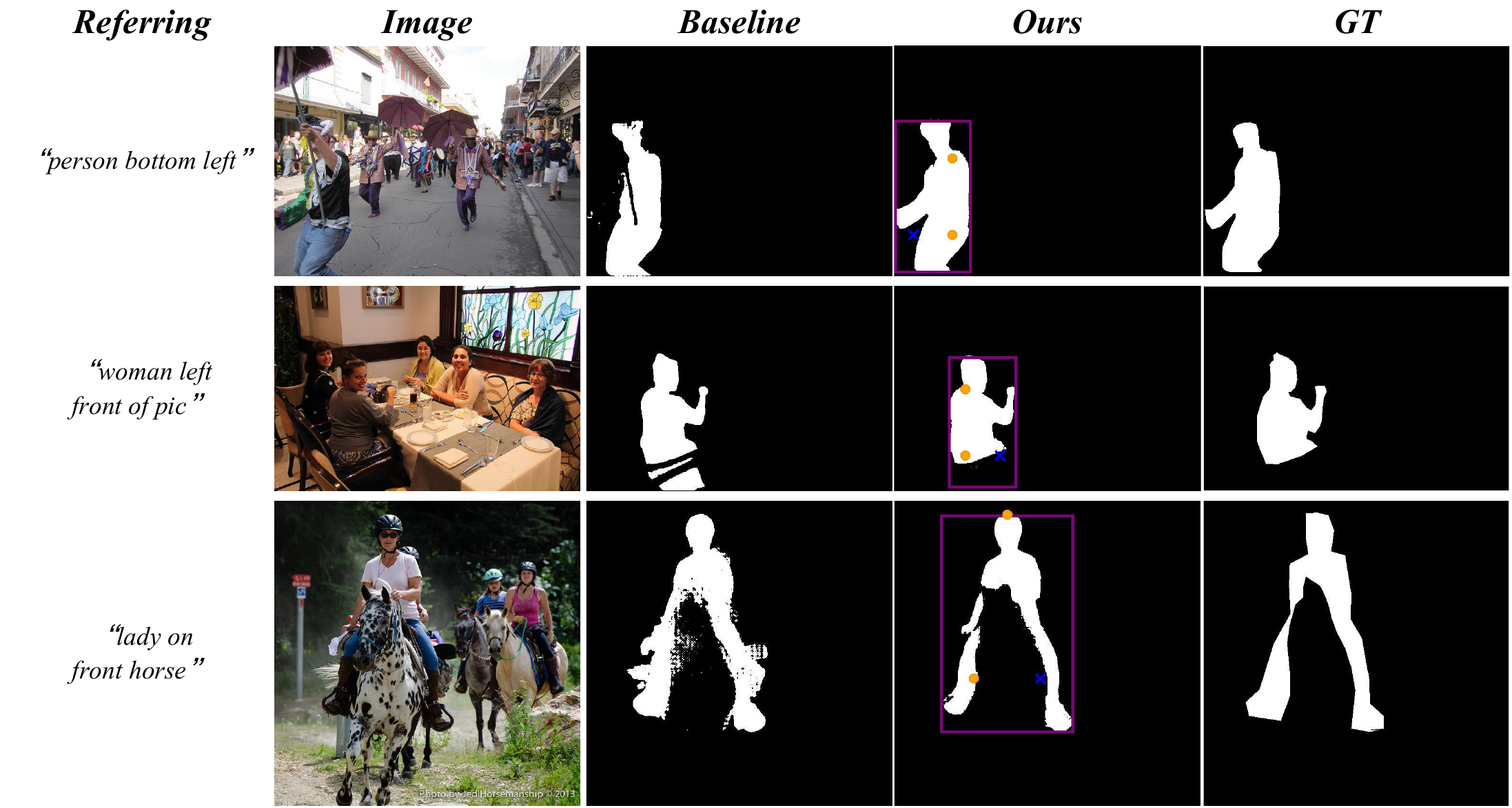}

   \caption{Visualization of qualitative results. Each row shows one RES example, with subfigures: referring text, input image, baseline result, our \model result, and GT (from left to right). Examples cover diverse challenging scenarios, and our \model achieves more accurate segmentation consistent with GT. }
   \label{fig:visualization}
   \vspace{-3mm}
\end{figure}

\textbf{Qualitative Analysis.} Figure \ref{fig:visualization} presents qualitative results of three representative RES scenarios, comparing our \model with the baseline to intuitively validate the effectiveness of our point discovery and reasoning design. These examples cover common challenging scenarios in RES, including fine-grained edge segmentation, background clutter suppression, and ambiguous target boundaries—all aligned with the quantitative gains observed earlier.  

In the first case (first row), the baseline generates a roughly correct mask but fails to capture the target’s arm region (due to the lack of negative points to suppress background interference) and exhibits blurry edges (attributed to insufficient positive points for fine-grained guidance). Our \model, leveraging EBD-selected high-value points, accurately retains the arm region and refines edge details, resulting in a mask more consistent with the ground truth (GT). The second example (second row) illustrates the baseline’s vulnerability to coarse bounding boxes: it incorrectly includes the underlying chair in the segmentation due to inadequate background filtering. Our VBR-driven negative point validation effectively identifies and excludes the chair region, correcting the bounding box imprecision and producing a target-specific mask. For the third scenario (third row) with ambiguous boundaries, the baseline’s segmentation is chaotic at the target-background interface, while our model’s synergistic use of positive points (capturing key semantic regions) and negative points (suppressing irrelevant areas) clarifies the ambiguous boundaries and delivers a coherent mask.  

Notably, despite the style adaptation of our segmentation decoder, minor discrepancies remain between our results and the GT. Our model generates smoother, more natural edges (free of sharp corners), whereas the GT occasionally contains label inconsistencies. This leads to cases where our visually superior segmentation is penalized by evaluation metrics, as subtle differences in fine-grained details are treated as missegmentation—highlighting a potential gap between metric-based evaluation and human visual perception for RES tasks.

\section{Conclusion}
In this work, we propose Referring Expression Segmentation Agent (\model) to address the key bottlenecks of existing MLLM-based RES methods: inefficient point sampling and unreliable text-coordinate reasoning. \model integrates Entropy-Based Point Discovery (EBD) and Vision-Based Reasoning (VBR). Leveraging the MLLM-inferred bounding box as a prior, EBD prioritizes points by information entropy. VBR then uses visual markers and conducts reasoning via VQA queries, freeing MLLMs from coordinate-based reasoning limitations. Early stopping is applied once sufficient qualified points are collected, and the final mask is generated. State-of-the-art results on multiple popular RES datasets validate the effectiveness of \model. Our exploration offers a novel perspective for future RES research, underscoring MLLMs’ potential to advance segmentation via aligned vision-language reasoning.



{
    \small
    \bibliographystyle{ieeenat_fullname}
    \bibliography{main}
}


\clearpage
\setcounter{page}{1}
\maketitlesupplementary
\section{Details for Entropy-Based Discovery}
\label{sec:appendix-entropy}

This section expands upon the entropy-based point discovery mechanism introduced in Sec.~\ref{sec:entropy}. Since the exact cross-modal Shannon entropy of an MLLM is not directly accessible, the core intuition is to approximate its spatial behavior through a distance-aware uncertainty field defined over the MLLM-predicted bounding box $\mathrm{bbox}^{\prime}$. As detailed in the main paper, regions near the geometric center and those near the boundary correspond to distinct high-information regimes: the former reflects semantic certainty but positional ambiguity, while the latter reflects spatial proximity but semantic uncertainty (both mapping to $p(\mathrm{pt})\!\approx\!0.5$ in the logistic calibration of Eq.~\ref{eq:entropy_eq}). 

\begin{figure*}[t]
  \centering
   \includegraphics[width=\linewidth]{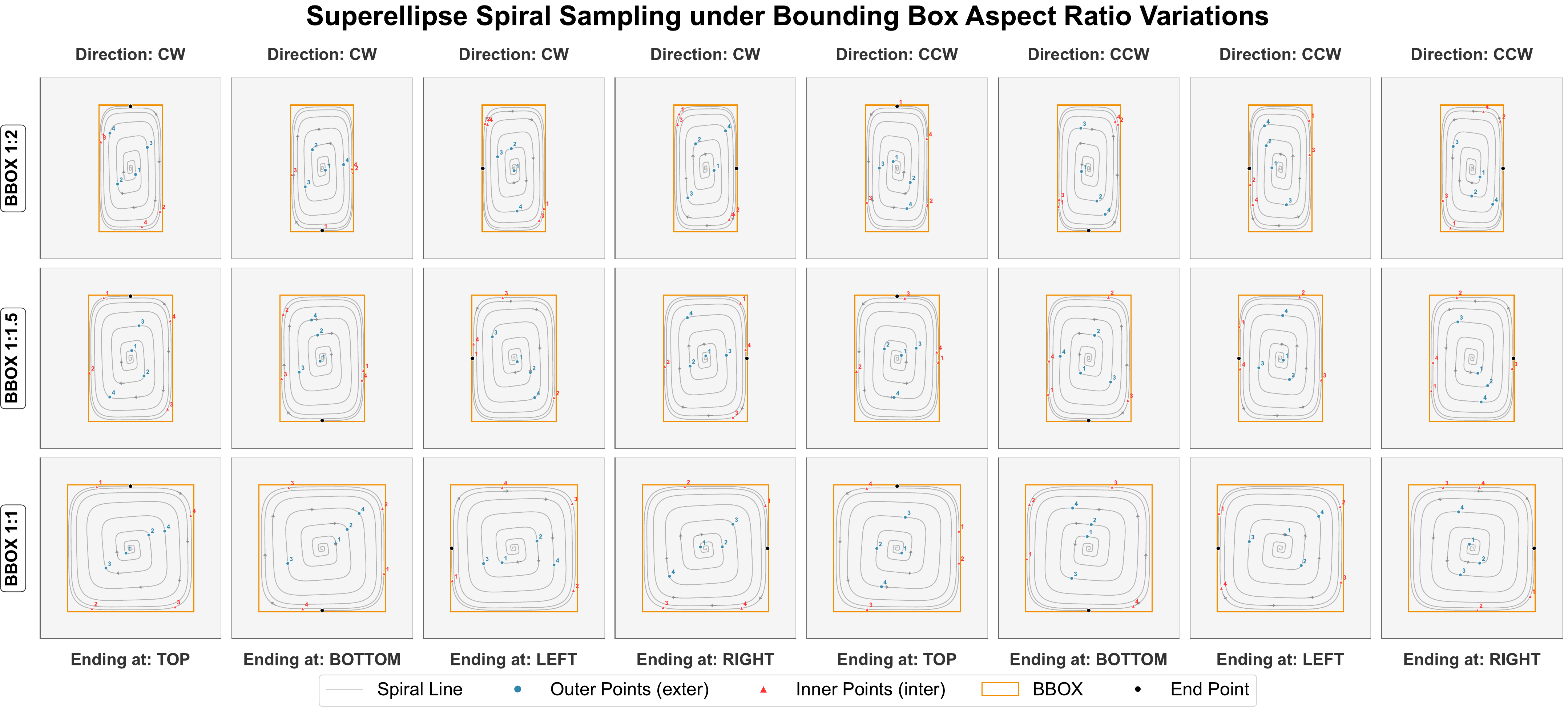}

   \caption{Visualization of the superellipse spiral sampling strategy across representative bounding box shapes. Each row corresponds to a different aspect ratio: elongated ($1\!:\!2$), moderately tall ($1\!:\!1.5$), and square ($1\!:\!1$) from top to bottom. Columns enumerate eight spiral configurations, defined by the combination of rotation direction (four clockwise followed by four counterclockwise) and terminal point (top, bottom, left, or right). hese variants demonstrate that the sampling density systematically concentrates in both center-proximal and boundary-proximal regions across all shapes and configurations. \textit{Note.} Aspect ratios such as $1.5\!:\!1$ or $2\!:\!1$ are omitted because they are equivalent up to rotation (e.g., a $2\!:\!1$ box is simply a $1\!:\!2$ box rotated by $90^\circ$), and the sampling procedure is designed to be orientation-invariant in normalized coordinates.
}
   \label{fig:supply_superellipse_sampling}
\end{figure*}

To operationalize this entropy-inspired perspective without computing $H(\mathrm{pt})$ explicitly, we instantiate a geometric proxy that approximates the structured uncertainty landscape via a parameterized superellipse spiral with adaptive sampling (Fig.~\ref{fig:framework}). The superellipse geometry encodes anisotropic proximity to edges (i.e., $\tilde{d}_e$), while a sigmoidal radial schedule follows the nonlinear transition in the entropy-calibrated probability space. An adaptive arc-length sampling mechanism further modulates point density based on the numerical gradient of normalized center distance—an analytically justified surrogate for $\nabla H$, thus ensuring dense coverage in high-uncertainty belts and sparse sampling elsewhere.

Finally, to reflect the two structural modes of uncertainty (center-proximal vs.\ boundary-proximal), sampled points are separated into two complementary candidate sets through a dual-queue strategy: the external queue scans from the bounding box edges inward, targeting boundary-proximal high-entropy regions, while the internal queue expands from the center outward to capture semantic interior ambiguity. This dual-queue design corresponds directly to our entropy landscape analysis and ensures that both informative regions are harvested before visual verification.

\subsection{Mathematical Formulation}
\subsubsection{Superellipse Spiral Generation}
The algorithm initiates by constructing a superellipse spiral within a predefined bounding box $\mathrm{bbox}^{\prime}$ (consistent with Sec.~\ref{sec:entropy}), which constrains the spatial extent of candidate points to the structured uncertainty field. Let the bounding box be defined as:
\[
\text{bbox}^{\prime} = [x_{\min}, y_{\min}, x_{\max}, y_{\max}]
\]
First, the center coordinates and semi-axes of the bounding box are computed to anchor the spiral—aligning with the distance metrics $d_c(\mathrm{pt})$ and $d_e(\mathrm{pt})$ in the entropy formulation:
\[
c_x = \frac{x_{\min} + x_{\max}}{2}, \quad c_y = \frac{y_{\min} + y_{\max}}{2}
\]
\[
a = \frac{x_{\max} - x_{\min}}{2}, \quad b = \frac{y_{\max} - y_{\min}}{2}
\]
where $c_x, c_y$ denote the center (the reference for $d_c(\mathrm{pt})$), and $a, b$ represent the semi-major and semi-minor axes (scaling the range of $d_e(\mathrm{pt})$).

The spiral is parameterized by the angle $\theta$ with $N_{\text{turns}}$ complete revolutions to ensure full coverage of the structured uncertainty field. For a total of $N_{\text{points}}$ points along the spiral path, the angle for the $i$-th point is:
\[
\theta_i = 2\pi \cdot N_{\text{turns}} \cdot \frac{i}{N_{\text{points}}-1}, \quad i = 0, 1, \dots, N_{\text{points}}-1
\]

To align with the sigmoidal calibration of $p(\mathrm{pt})$ in the entropy framework and achieve smooth radial growth (avoiding abrupt acceleration or deceleration), the radius follows a sigmoidal function:
\[
r(t) = \frac{1}{1 + \exp(-k_{\text{sigmoid}}(t - t_0))}, \quad t = \frac{i}{N_{\text{points}}-1}
\]
where $t \in [0,1]$ normalizes the index $i$ to the spiral's full extent, $k_{\text{sigmoid}}$ denotes the steepness coefficient of the sigmoid function (mirroring the mapping in Eq. \ref{eq:entropy_eq}), and $t_0$ denotes the midpoint offset of the sigmoid curve.

A superellipse transformation with exponent $n$ is applied to shape the spiral, enhancing adherence to the bounding box edges (relevant for $d_e(\mathrm{pt})$) compared to a standard ellipse. The transformation proceeds in four steps for each $\theta_i$:

1. \textbf{Preliminary ellipse coordinates}: Compute base ellipse points using the radial growth function, encoding the distance-dependent uncertainty:
\[
x_{\text{ellipse}} = a \cdot r(t) \cdot \cos(\theta_i), \quad y_{\text{ellipse}} = b \cdot r(t) \cdot \sin(\theta_i)
\]

2. \textbf{Superellipse normalization}: Normalize coordinates to the unit superellipse, aligning with the normalized distance metrics $\tilde{d}_c, \tilde{d}_e \in [0,1]$:
\[
x_{\text{norm}} = \left\lvert \frac{x_{\text{ellipse}}}{a} \right\rvert, \quad y_{\text{norm}} = \left\lvert \frac{y_{\text{ellipse}}}{b} \right\rvert
\]
\[
\phi_{norm} = \left(x_{\text{norm}}^n + y_{\text{norm}}^n\right)^{1/n}
\]

3. \textbf{Scale to superellipse boundary}: Adjust coordinates to lie exactly on the superellipse contour, ensuring precise mapping to $d_e(\mathrm{pt})$:
\[
\alpha_{scale} = \frac{r(t)}{\phi_{norm}}
\]
\[
x_{\text{super}} = x_{\text{ellipse}} \cdot \alpha_{scale}, \quad y_{\text{super}} = y_{\text{ellipse}} \cdot \alpha_{scale}
\]

4. \textbf{Translate to final coordinates}: Shift points to the bounding box center, anchoring the spiral to the reference of $d_c(\mathrm{pt})$:
\[
x_i = c_x + x_{\text{super}}, \quad y_i = c_y + y_{\text{super}}
\]

To introduce configuration variability while preserving the structured uncertainty, the spiral direction (clockwise or counterclockwise) and endpoint positions are randomly selected from predefined valid combinations.

\subsubsection{Adaptive Arc-Length Sampling}
The sampling process employs adaptive arc-length adjustment to optimize point density—prioritizing regions where $H(\mathrm{pt})$ is maximized (i.e., $p(\mathrm{pt}) \approx 0.5$) by increasing density in regions of rapid radial growth (balancing $d_c$ and $d_e$) and maintaining sparser coverage in uniform uncertainty regions.

First, calculate the cumulative arc-length of the spiral: Let $\mathbf{p}_i = (x_i, y_i)$ denote the $i$-th spiral point. The Euclidean distance between consecutive points (segment length) is:
\[
\Delta s_i = \lVert \mathbf{p}_{i+1} - \mathbf{p}_i \rVert_2 = \sqrt{(x_{i+1} - x_i)^2 + (y_{i+1} - y_i)^2}
\]
The cumulative arc-length up to the $i$-th point is:
\[
S_i = \sum_{j=0}^{i-1} \Delta s_j, \quad S_0 = 0
\]
where the total arc-length of the spiral is $S_{\text{total}} = S_{N_{\text{points}}-1}$.

The core innovation of the sampling process lies in dynamic step size adjustment, driven by the radial growth rate (a surrogate for the entropy gradient), which proceeds as follows:

1. \textbf{Radial distance computation}: Calculate the radial distance of each spiral point from the center, directly corresponding to $\tilde{d}_c(\mathrm{pt})$:
\[
d_i = \sqrt{(x_i - c_x)^2 + (y_i - c_y)^2}
\]

2. \textbf{Normalized radial distance}: Normalize $d_i$ to the range $[0,1]$, matching the normalization of $\tilde{d}_c$ in the entropy formulation:
\[
d_{\text{norm},i} = \frac{d_i - \min\limits_j(d_j)}{\max\limits_j(d_j) - \min\limits_j(d_j)}
\]

3. \textbf{Radial growth rate}: Estimate the numerical gradient of $d_{\text{norm},i}$ to capture the rate of change in $\tilde{d}_c$, a proxy for the entropy gradient $\nabla H(\mathrm{pt})$:
\[
g_i = \frac{d}{dt}d_{\text{norm},i} \approx \frac{d_{\text{norm},i+1} - d_{\text{norm},i-1}}{2\Delta t}
\]
where $\Delta t = 1/(N_{\text{points}}-1)$ is the step size in the normalized index $t$.

\subsubsection{Entropy Gradient Proxy Justification}
To explicitly link $g_i$ to $\nabla H(\mathrm{pt})$, we establish a three-step correlation chain:

\begin{itemize}
\item \textbf{Variable Correspondence}: $d_{\text{norm},i}$ directly maps to $\tilde{d}_c(\mathbf{p}_i)$ (Sec.~\ref{sec:entropy}), as both represent normalized center distance for point $\mathbf{p}_i$.

\item \textbf{Correlation between $g_i$ and $dp/dt$}: From the logistic mapping $p(\mathbf{p}_i) = \sigma(a - b \cdot d_{\text{norm},i} + c \cdot \tilde{d}_e(\mathbf{p}_i))$, the derivative is $dp/dt = \sigma'(\cdot) \cdot (-b \cdot g_i)$. Since $\sigma'(z) = \sigma(z)(1-\sigma(z)) > 0$ and $b > 0$, $|dp/dt|$ is proportional to $|g_i|$ (up to a positive constant).

\item \textbf{Correlation between $|dp/dt|$ and $|\nabla H(\mathrm{pt})|$}: For Shannon entropy $H(p) = -p\log p - (1-p)\log(1-p)$, $dH/dt = -\left(\log\frac{p}{1-p}\right) \cdot dp/dt$. Near $p \approx 0.5$ (high-entropy regions), $\left|\log\frac{p}{1-p}\right|$ is large and smooth, so $|\nabla H(\mathbf{p}_i)|$ is positively correlated with $|dp/dt|$.
\end{itemize}

Combining these, $|g_i|$ is positively correlated with $|\nabla H(\mathbf{p}_i)|$, validating that step size adjustment via $g_i$ prioritizes high-information-gain regions.

4. \textbf{Normalized growth rate}: Scale $g_i$ to $[0,1]$ for consistent coefficient mapping, aligning with the calibrated probability space of $p(\mathrm{pt})$:
\[
g_{\text{norm},i} = \frac{g_i - \min\limits_j(g_j)}{\max\limits_j(g_j) - \min\limits_j(g_j)}
\]

5. \textbf{Dynamic coefficient calculation}: Map the normalized growth rate to a step size adjustment coefficient within the range $[k_{\text{min}}, k_{\text{max}}]$:
\[
k_i = k_{\text{min}} + (k_{\text{max}} - k_{\text{min}}) \cdot g_{\text{norm},i}
\]
where $k_i < (k_{\text{min}} + k_{\text{max}})/2$ reduces step size (increases density) for regions of rapid radial growth (high entropy, $p(\mathrm{pt}) \approx 0.5$), and $k_i > (k_{\text{min}} + k_{\text{max}})/2$ increases step size (reduces density) for regions of slow radial growth (low entropy, $p(\mathrm{pt}) \approx 0$ or $1$).

With the dynamic coefficient defined, the sampling point generation proceeds iteratively with controlled randomness—mimicking the greedy ranking of $H(\mathrm{pt})$ in the dual-queue strategy:

1. \textbf{Base step size definition}: Define the base step as the maximum dimension of the bounding box, scaling with the range of $\tilde{d}_e(\mathrm{pt})$:
\[
\beta = \max(x_{\max} - x_{\min}, y_{\max} - y_{\min})
\]

2. \textbf{Initial bias introduction}: Introduce a random initial offset to avoid deterministic sampling, ensuring diversity in high-entropy candidate selection:
\[
b \sim \mathcal{U}(0, \beta)
\]
where $\mathcal{U}(a,b)$ denotes the uniform distribution over the interval $[a,b]$.

3. \textbf{Iterative sampling distance generation}: Generate sampling positions along the cumulative arc-length:
\[
s_0 = b
\]
\[
s_{j+1} = s_j + \beta \cdot k_{I(s_j)}
\]
where $I(s_j)$ is the index of the spiral point closest to the arc-length $s_j$.

4. \textbf{Random perturbation application}: Add controlled noise to enhance variability while preserving spatial coherence—mirroring the stochasticity in ranking high-entropy points:
\[
\delta_j \sim \mathcal{U}(-\epsilon \cdot \beta, \epsilon \cdot \beta)
\]
\[
s'_j = s_j + \delta_j
\]
with boundary constraints $s'_0 \geq 0$ and $s'_{\text{final}} \leq S_{\text{total}}$ to ensure sampling stays within the structured uncertainty field of $\mathrm{bbox}^{\prime}$.

5. \textbf{Linear interpolation for final points}: Compute the final sampling points by linearly interpolating between consecutive spiral points along the adjusted arc-length $s'_j$, yielding candidates that densely populate high-entropy regions.

\subsubsection{Internal and External Point Separation}
To directly instantiate the dual-queue greedy discovery strategy in Sec.~\ref{sec:entropy}, the sampled points are partitioned into two distinct candidate sets that align with the edge-inward and center-outward scanning logic:

\begin{itemize}
\item \textbf{External points}: The first $\mathcal{K}$ points in the sampling sequence, concentrated in the spiral's outer regions to harvest boundary-proximal high-entropy candidates (matching the internal queue in Sec.~\ref{sec:entropy}).

\item \textbf{Internal points}: The first $\mathcal{K}$ points from the reversed sampling sequence, concentrated in the spiral's core area to harvest center-proximal high-entropy candidates (matching the external queue in Sec.~\ref{sec:entropy}). 
\end{itemize}

This separation strategy ensures balanced coverage of the structured uncertainty field, with each candidate set targeting the specific high-information regions defined by the entropy-based framework, and retains the top-$\mathcal{K}$ candidates (per the candidate budget) for subsequent verification.

\subsection{Parameter Configuration}
The algorithm employs the following fixed parameter settings to ensure alignment with the entropy-based framework and robust performance across diverse application scenarios. All parameters are explicitly calibrated to the entropy model (Sec.~\ref{sec:entropy}) for cross-module consistency:

\textbf{Spiral Generation Parameters}

\begin{itemize}
    \item Number of turns: $N_{\text{turns}} = 8$ (ensures full coverage of the structured uncertainty field)
    \item Total number of spiral points: $N_{\text{points}} = 3000$ (balances sampling resolution and computational efficiency)
    \item Superellipse exponent: $n = 5.0$ (calibrated to $\tilde{d}_e(\mathrm{pt})$—enhances boundary adherence to cover edge-proximal high-entropy regions)
    \item Sigmoid steepness coefficient: $k_{\text{sigmoid}} = 8$ (aligned with logistic mapping weights $(b,c)$—matches the steepness of $p(\mathrm{pt})$'s nonlinear transition)
    \item Sigmoid midpoint offset: $t_0 = 0.5$ (coincides with $p(\mathrm{pt}) \approx 0.5$ peak entropy region)
    \item Radial growth function: Sigmoidal (defined as $r(t) = \frac{1}{1 + \exp(-k_{\text{sigmoid}}(t - t_0))}$)
\end{itemize}

\textbf{Adaptive Sampling Parameters}
\begin{itemize}
    \item Dynamic coefficient range: $[k_{\text{min}}, k_{\text{max}}] = [0.5, 1.5]$ (calibrated to $|\nabla H(\mathrm{pt})|$—reduces step size for large entropy gradients, increases for small ones)
    \item Random perturbation factor: $\epsilon = 0.2$ (20\% of base step, mirrors entropy ranking stochasticity to avoid local optima)
    \item Base step size: $\max(\text{bbox width}, \text{bbox height})$ (scales with $\tilde{d}_e(\mathrm{pt})$'s range)
    \item Number of samples per candidate set: $\mathcal{K}$ (aligned with the candidate budget in Sec.~\ref{sec:entropy})
\end{itemize}

\subsection{Visualization of Spiral Sampling Behavior}
\label{sec:spiral-visualization}

Fig.~\ref{fig:supply_superellipse_sampling} illustrates the behavior of the proposed entropy-inspired spiral sampling method under different bounding box geometries and spiral configurations. Because the sampling strategy operates entirely in normalized coordinate space within $\mathrm{bbox}^{\prime}$, its behavior is inherently invariant to absolute orientation, scale, or image content.

To validate this property, we visualize the sampling trajectories over bounding boxes with three representative aspect ratios ($1\!:\!1$, $1\!:\!1.5$, and $1\!:\!2$), arranged from bottom to top. Within each row, the eight columns enumerate the Cartesian product of two rotation directions (clockwise (CW) vs.\ counterclockwise (CCW) ) and four terminal points (top, bottom, left, or right).

These visualizations confirm three key properties of our strategy:  

\textbf{(1) Geometry-consistent exploration.} Regardless of rotation direction or aspect ratio, sampling density is automatically concentrated in two information-rich zones—center-proximal and boundary-proximal regions—matching the entropy landscape analyzed in Sec.~\ref{sec:entropy}. 

\textbf{(2) Parameter-invariant coverage.} Both rotation direction and endpoint influence only the traversal order, not the spatial distribution of candidate points; the resulting coverage remains stable across configurations.  

\textbf{(3) Shape-adaptability.} The superellipse transformation allows the spiral to adhere naturally to different bounding box shapes without manual heuristics or grid redesign. Even elongated boxes (e.g., $1\!:\!2$) retain balanced sampling across the structured uncertainty field, highlighting the robustness of the geometric proxy.

Together, these results demonstrate that the proposed spiral sampling strategy is not only flexible and parameter-efficient, but also faithful to the entropy-based intuition that informed its design.



\section{Further Analysis}
\label{label:further_analysis}

\subsection{Oracle Experiments}
\label{sec:supply_oracle_bbox_prior}

\begin{table*}[t]
    \centering
    \small 
    \begin{tabularx}{\textwidth}{Xcccccc} 
        \toprule
        \multirow{2}{*}{\bf Setting} & \multirow{2}{*}{\bf Sampling} & \multicolumn{3}{c}{\bf RefCOCO } & \multirow{2}{*}{\bf Avg.} \\
        \cmidrule(lr){3-5}
        & & \bf Val & \bf TestA & \bf TestB & \\
        \midrule
        \multirow{2}{*}{Gold bbox (tight)}  & Random (pos2+neg1)  & 88.57 & 88.09 & 88.84 & 88.50 \\
                                & \cellcolor{blue!7.5}Spiral (pos2+neg1) & \cellcolor{blue!7.5} 88.70 & \cellcolor{blue!7.5} \bf 89.07 &  \cellcolor{blue!7.5} 89.30 & \cellcolor{blue!7.5} \bf89.02 \\
        \midrule
        \multirow{2}{*}{Gold bbox (mild $10\%$, one-side) } & Random (pos2+neg1)  & 88.50 \textcolor{red}{(-0.07)} & 88.02 \textcolor{red}{(-0.07)} & 89.11 \textcolor{red}{(+0.27)} & 88.54 \textcolor{red}{(+0.04)} \\
                               &\cellcolor{blue!7.5} Spiral (pos2+neg1) & \cellcolor{blue!7.5} \bf88.83 \textcolor{blue}{(+0.13)} &  \cellcolor{blue!7.5} 88.29 \textcolor{blue}{(-0.78)} &  \cellcolor{blue!7.5} \bf89.31 \textcolor{blue}{(+0.01)} & \cellcolor{blue!7.5} \bf88.81 \textcolor{blue}{(-0.21)} \\
        \midrule
        \multirow{2}{*}{Gold bbox (severe $5\%-15\%$, per-side)} & Random (pos2+neg1)  & 86.16 \textcolor{red}{(-2.41)} & 85.79 \textcolor{red}{(-2.30)} & 86.65 \textcolor{red}{(-2.19)} & 86.20 \textcolor{red}{(-2.30)} \\
                               &\cellcolor{blue!7.5} Spiral (pos2+neg1) & \cellcolor{blue!7.5} 87.25 \textcolor{blue}{(-1.45)} &  \cellcolor{blue!7.5} 87.19 \textcolor{blue}{(-1.88)} &  \cellcolor{blue!7.5} 87.56 \textcolor{blue}{(-1.74)} & \cellcolor{blue!7.5} 87.33 \textcolor{blue}{(-1.69)} \\
        \bottomrule
    \end{tabularx}
    \caption{Comparison of random sampling vs. our superellipse-spiral sampling under bbox perturbations on RefCOCO splits. {\color{red}Red}/{\color{blue}blue} annotations denote mIoU change relative to the same sampling strategy’s gold bbox (tight) performance ({\color{red}Red}: Random, {\color{blue}blue}: Spiral).}
    \label{tab:suppl_oracle_perturb_compare}
\end{table*}

\subsubsection{Gold Bbox Oracle Experiments}

A key advantage of our superellipse-spiral sampling (over conventional random probing) is its enhanced robustness to bounding box (bbox) perturbations—critical for simulating real-world localization noise. This section evaluates the upper bound of both sampling strategies under controlled bbox perturbations, with results summarized in Table~\ref{tab:suppl_oracle_perturb_compare}.

We define two perturbation regimes to mimic practical localization inaccuracies:

1. \textbf{+$10\%$ one-side expansion}: Randomly expand the gold bbox (directly derived from annotations) by 10\% along one arbitrary direction (e.g., left, top).

2. \textbf{+$[5-15\%]$ per-side expansion}: Independently expand each of the four bbox sides by a random factor between 5\% and 15\%—a more severe perturbation that better reflects real-world bbox noise.

\paragraph{Performance Robustness Under Perturbations.}

For each strategy, we quantify performance degradation as the difference between perturbed and gold bbox (tight) mIoU (marked in red for Random, blue for Spiral in Table~\ref{tab:suppl_oracle_perturb_compare}).

- Under mild $+10\%$ one-side expansion: Random sampling shows negligible changes (average $±0.04$ mIoU), while our Spiral sampling maintains stable performance (average $-0.21$ mIoU). Both strategies perform well here due to the limited perturbation scope.

- Under severe $+[5-15\%]$ per-side expansion: Random sampling suffers a substantial drop of 2.30 average mIoU, whereas our Spiral sampling only degrades by 1.69 average mIoU.

Across all RefCOCO splits (Val, TestA, TestB), Spiral sampling outperforms Random sampling under both perturbation regimes, with the gap widening in the severe noise scenario. Across all three splits, spiral sampling reduces the mIoU drop by an average of 0.6 mIoU under mild bbox perturbation and 0.61 mIoU under severe perturbation, compared to random sampling. This confirms that our geometry-aware design is inherently resilient to bbox localization errors.

\paragraph{Why spiral sampling is more robust.}
The superior robustness stems from two core design choices:

1. The spiral’s sigmoidal radial schedule concentrates probes in center-proximal regions, which are less likely to drift outside the true mask even with moderate bbox shifts.

2. The superellipse shape adapts to the bbox aspect ratio, ensuring boundary-aligned probes remain informative despite slight perturbations.

These properties eliminate the need for expensive bbox refinement or stronger supervision, offering a practical solution to reduce sensitivity to localization noise.

We note that the performance drop observed under severe perturbation is comparable to the gap between predicted and gold bounding boxes in typical RES benchmarks, underscoring the need for more robust geometric sampling strategies in real-world deployments.

\subsection{Upper Bound Results}
To quantify the performance headroom of the Referring Expression Segmentation (RES) task independent of model design, we conduct oracle experiments. We replace predicted bounding boxes (bboxes), predicted point prompts, or both with ground-truth (gold) signals, with results summarized in Table~\ref{tab:supply_upper_bound}.

\begin{table}[htbp]
    \centering
    \setlength{\tabcolsep}{3pt}
    \small
    \begin{tabular}{lcccc}
        \toprule
        \multirow{2}{*}{\bf BBox + Points} & \multicolumn{3}{c}{\bf RefCOCO} & \multirow{2}{*}{\bf Avg.} \\
        \cmidrule(lr){2-4}
        & \bf Val & \bf TestA & \bf TestB & \\
        \midrule
        \rowcolor{gray!10}Pred + Pred (Reported) & 81.19 & 82.64 & 78.07 & 80.63 \\
         Pred  + Gold  & \bf 83.16 & \bf 83.31 & \bf 80.76& \bf 82.41 \\
         Gold + Pred & 86.28 & 86.52 & 87.15&  86.65 \\
         \rowcolor{blue!7.5}\bf Gold  + Gold (Upper Bound) & \bf 88.70 & \bf 89.07 & \bf 89.30& \bf 89.02 \\
        \bottomrule
    \end{tabular}
    \caption{Upper bound results of RES task under different oracle settings. "Pred" denotes model-predicted signals, and "Gold" denotes ground-truth signals.}
    \label{tab:supply_upper_bound}
\end{table}

\paragraph{Oracle Upper Bounds Analysis}

- \textbf{Pred+Gold vs. Pred+Pred}: Replacing learned points with gold annotations yields only a modest gain of 1.8 mIoU, suggesting our sampling effectively captures high-value points even under imperfect bboxes.

- \textbf{Gold+Pred vs. Pred+Pred}: Substituting predicted bboxes with gold bboxes (while retaining automatic point reasoning) brings a larger boost of ~6.0 mIoU (86.65 vs. 80.63). This underscores bbox localization as the dominant error source in the current pipeline.

- \textbf{Gold+Gold (Upper Bound)}: The full oracle setting achieves an average mIoU of 89.02, setting the empirical upper bound. This shows further improvements will largely depend on enhancing spatial grounding before point refinement.

\paragraph{Rationale for Not Training a Bbox Regression Module}

The oracle results reveal an 8.4 mIoU gap between the reported (Pred+Pred) and upper-bound (Gold+Gold) settings, with bbox localization being the key bottleneck. However, we intentionally avoid training a dedicated bbox regressor for two core reasons:

1. Our work focuses on evaluating whether a lightweight point-based agent can compensate for coarse localization without full supervision or detection-specific training. The small gap between Pred+Pred and Pred+Gold confirms the proposed agent already captures high-quality spatial cues under imperfect bboxes.

2. Bbox prediction is a plug-in module that can be replaced by advanced detectors (e.g., GroundingDINO, multimodal vision transformers for region proposals). Integrating such detectors is orthogonal to our core contribution and left as future work.

These findings highlight a promising direction: combining a high-quality pretrained detection backbone with our geometry-guided point reasoning agent to bridge the gap toward the gold oracle. Overall, our analysis indicates that integrating a stronger localization module is both orthogonal to and highly complementary with our agent-based reasoning pipeline—a promising avenue for future RES systems. 

\subsection{On Special Tokens for Numeric Coordinates}
While recent works have explored specialized tokenization or number embeddings to better encode numeric coordinates in language models, such textualized representations remain fundamentally disconnected from image-space geometry. Text tokens---whether raw digits or custom markers---do not encode spatial continuity or visual context, and current MLLMs are not pretrained to ground such symbolic inputs in pixel space.

Token-based numeric embeddings, though effective for text-only arithmetic tasks, fail to capture fundamentally different geometric priors required for vision tasks: bounding boxes and point prompts differ not only in meaning but also in spatial semantics. Introducing dedicated tokens for both cases either leads to token explosion or requires extensive retraining to achieve a vague form of grounding.

In contrast, our agent-based visual reasoning strategy bypasses the need for textual abstraction entirely. We operate directly in the pixel domain, constructing high-information probe points based on geometric uncertainty and spatial reasoning. This approach yields context-aware, semantics-aligned point discovery without modifying the MLLM vocabulary or training objective. It also remains compatible with arbitrary MLLM backbones and domain settings, providing a simple yet effective alternative to specialized token schemes.

\subsection{Point Reasoning Analysis under Zero-Shot Settings}
\label{sec:reasoning-analysis}

\begin{figure}[htbp]
  \centering
   \includegraphics[width=\linewidth]{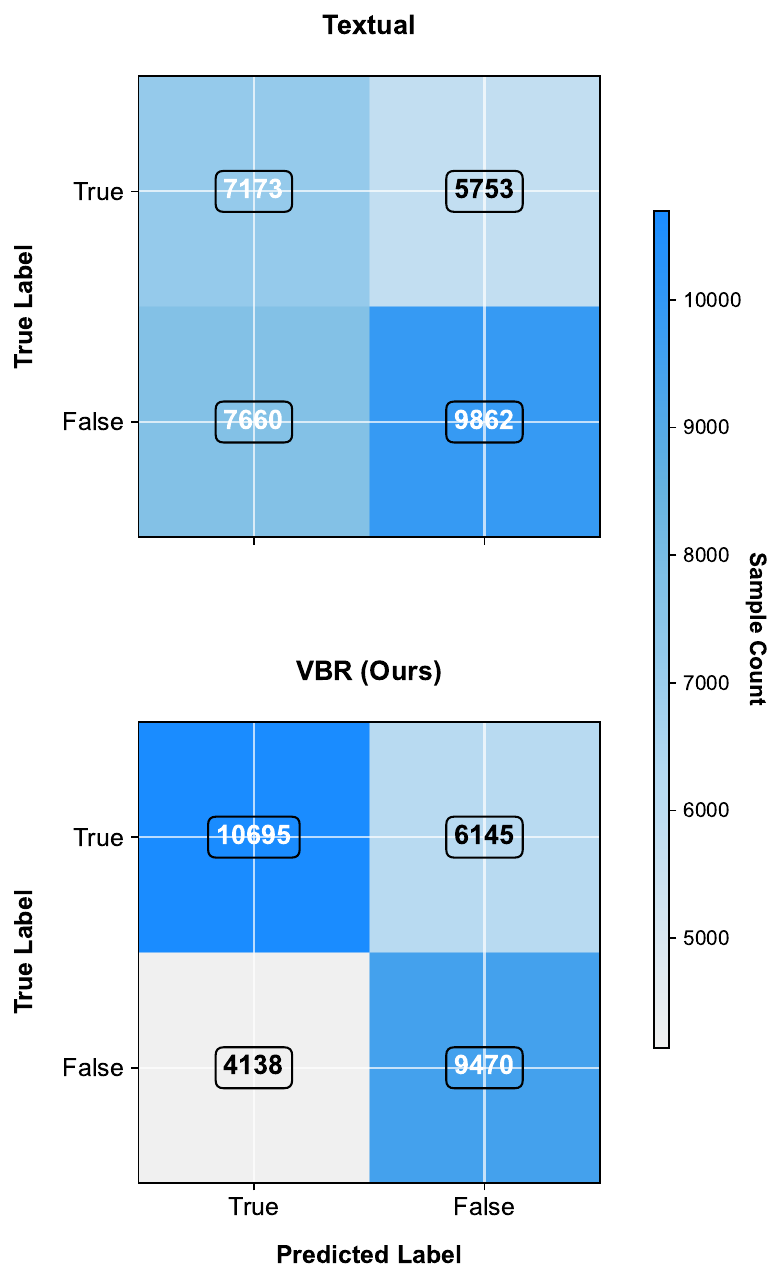}
   \caption{Comparison of confusion matrices for textual coordinate reasoning (`\texttt{(x,y)}') vs.\ VBR, under zeroshot validation on RefCOCO val split. VBR significantly increases true positives while reducing false positive and false negative errors.}
   \label{fig:supply_confusion_matrix}
\end{figure}

This section evaluates the reliability of point reasoning when expressed either as textualized coordinate prompts or via our geometry-aware visual behavior reasoning (VBR) module. Across all evaluation conditions—prompt format variations, confidence thresholds, and spatial regions—VBR consistently outperforms textualized coordinate reasoning by large margins. This confirms that our approach mitigates the inherent brittleness of token-based numeric inputs, providing stable and spatially grounded reasoning that scales across different prompt styles. Full experimental evidence is summarized in Table~\ref{tab:reasoning_format_comparison}, Table~\ref{tab:threshold_conf}, and Fig.~\ref{fig:supply_confusion_matrix}.

\paragraph{Reasoning Template.}
We use the following unified template for all point reasoning experiments (both textualized and visual), where the format of point expression is substituted into \texttt{\{FORMAT\}} and the referring text is substituted into $\mathcal{T}$:

\begin{quote}
\emph{``Answer strictly yes or no: Is the Point \{FORMAT\} on the object referred to by `\{$\mathcal{T}$\}' in the picture?''}
\end{quote}

\paragraph{Thresholding Policy for Zero-Shot Reasoning.}
Both textualized coordinate reasoning and our VBR are evaluated in a strictly zero-shot setting (no task-specific training).
Because VBR produces continuous confidence scores, we apply a simple thresholding step at inference to convert them into binary decisions; this operation is \emph{not} a form of training. To ensure fairness, all methods are evaluated under consistent thresholds ($\eta = 0.0$, $0.6$, $0.7$, and $0.8$), and $\eta = 0.6$ is used as the default for main comparison tables. We also report threshold-sensitivity results in Appendix Table~\ref{tab:threshold_conf}, and provide F1/precision-recall curves in the supplementary materials to demonstrate robustness to threshold choice.

\paragraph{Effect of Prompt Format.}
Table~\ref{tab:reasoning_format_comparison} compares different textualized coordinate formats against VBR, under zero thresholding ($\eta = 0.0$) to retain all points. While textual formats struggle to achieve a balanced precision-recall trade-off due to the symbolic nature of numeric tokens, VBR consistently delivers significantly higher performance across all metrics without requiring explicit numeric token handling.

\begin{table}[htbp]
    \centering
    \small
    \setlength{\tabcolsep}{3pt}
    \caption{Effect of coordinate prompt format on zero-shot point reasoning ({\bf RefCOCO val}), with identical sampling and bbox priors ($\eta = 0.0$).}
    \label{tab:reasoning_format_comparison}
    \begin{tabular}{lcccc}
        \toprule
        \bf Reasoning Format & \bf $\text{Acc}\uparrow$ & \bf $\text{Precision}\uparrow$ & \bf $\text{Recall}\uparrow$ & \bf $\text{F1}\uparrow$ \\
        \midrule
        \textit{Textual} `\texttt{(x,y)}'         & 55.95 & 55.49 & 48.36 & 51.68 \\
        \textit{Textual} `\texttt{x is, y is}'    & 54.87 & 53.98 & 49.95 & 51.89 \\
        \textit{Textual} `\texttt{x=, y=}'        & 53.78 & 53.15 & 43.11 & 47.61 \\
        \rowcolor{blue!7.5} \textbf{VBR (ours)}   & \textbf{66.23} & \textbf{63.51} & \textbf{72.10} & \textbf{67.53} \\
        \bottomrule
    \end{tabular}
\end{table}

\paragraph{Error Profile Comparison.}
Figure~\ref{fig:supply_confusion_matrix} visualizes the confusion matrices of the simplest textual coordinate format (`\texttt{(x,y)}') versus our VBR under zero-shot validation. VBR produces significantly fewer false positives and false negatives, with a large increase in true positives (TP), confirming that VBR reasoning is not only more accurate but also more \emph{discriminative} and \emph{spatially grounded}.

\subsection{Threshold Sensitivity and Point Availability in Zero-Shot Reasoning}

Table~\ref{tab:threshold_conf} reports the performance of textualized coordinate reasoning versus our VBR across increasing confidence thresholds $\eta$.
Both methods are evaluated in a strictly zero-shot setting without any fine-tuning, and we observe two important trends.

First, VBR demonstrates consistently higher accuracy, precision, recall, and F1 across all thresholds.
At a moderate threshold ($\eta=0.6$), VBR improves F1 by +13.4 points, showing that its soft confidence outputs are better aligned with true membership probabilities than textualized outputs processed as discrete tokens.

Second, thresholding rapidly reduces the number of valid points for textualized coordinates—in particular, only 1.72 out of 8 sampled points remain at $\eta=0.8$, making the prompt highly unstable even when remaining points are correct.
In contrast, VBR retains most points even at high confidence ($\eta=0.8$), preserving an average of 5.28 query points while maintaining superior quality.
This confirms that VBR not only improves point-wise reasoning accuracy but also provides a more robust confidence landscape for downstream refinement.

Together, these results highlight that the proposed visual behavior reasoning is not only more accurate but also produces better-calibrated uncertainty estimates, preserving candidate diversity and reducing the risk of empty prompt failures in zero-shot RES scenarios.

\begin{table}[t]
    \centering
    \setlength{\tabcolsep}{3pt} 
    \small
    \begin{tabular}{l|c|cccc|c}
        \toprule
        \bf $\eta=$ & \bf Method & \bf $\text{Acc}\uparrow$ & \bf $\text{Precision}\uparrow$ & \bf $\text{Recall}\uparrow$ & \bf $\text{F1}\uparrow$ & Available\\
        \midrule
        \multirow{2}{*}{0}  & Textual  & 55.95 & 55.49 & 48.36 & 51.68 & 7.99\\
        
        &\cellcolor{blue!7.5} \bf VBR & \cellcolor{blue!7.5}  \bf66.23 &  \cellcolor{blue!7.5} \bf63.51 &  \cellcolor{blue!7.5} \bf 72.10 & \cellcolor{blue!7.5} \bf 67.53 & \cellcolor{blue!7.5} 7.99 \\
        \midrule
                                
        \multirow{2}{*}{0.6} & Textual  & 57.71 & 58.36 & 54.78 & 56.51 & 5.28 \\
        
        & \cellcolor{blue!7.5}\bf VBR & \cellcolor{blue!7.5} \bf 67.46 & \cellcolor{blue!7.5} \bf 64.95 &  \cellcolor{blue!7.5} \bf75.78 & \cellcolor{blue!7.5} \bf 69.95 & \cellcolor{blue!7.5} \bf 7.05 \\
        \midrule

        \multirow{2}{*}{0.7} & Textual  & 59.56 & 61.66 & 59.26 & 60.43 & 3.34 \\
        
        &\cellcolor{blue!7.5} \bf VBR & \cellcolor{blue!7.5}  \bf68.77 &  \cellcolor{blue!7.5} \bf66.56 &  \cellcolor{blue!7.5} \bf 78.34 & \cellcolor{blue!7.5} \bf71.97 & \cellcolor{blue!7.5} \bf 6.23\\
        \midrule
        
        \multirow{2}{*}{0.8} & Textual  & 61.77 & 64.63 & 66.37 & 65.49 & 1.72\\
        &\cellcolor{blue!7.5}\bf VBR & \cellcolor{blue!7.5}  \bf70.16 &  \cellcolor{blue!7.5} \bf68.13 &  \cellcolor{blue!7.5} \bf 81.62 & \cellcolor{blue!7.5} \bf74.27 & \cellcolor{blue!7.5} \bf 5.28\\
        \bottomrule
    \end{tabular}
    \caption{
    Comparison of textualized coordinate reasoning and our VBR under different confidence thresholds $\eta$ (higher $\eta$ is stricter).
    VBR consistently outperforms text-based prompting across all metrics, achieving higher precision-recall balance (F1) while retaining significantly more usable points after filtering.
    This indicates that VBR produces more reliable and geometrically grounded confidence scores, making it suitable for threshold-based refinement in zero-shot setups.
    }

    \label{tab:threshold_conf}
\end{table}

\subsection{Supplementary Results with Different Base Models}

\begin{table*}[htbp]
\small
\centering
\setlength{\tabcolsep}{3pt}
\caption{Comparison with state-of-the-art methods on image referring expression segmentation (RES) and reasoning segmentation datasets-RefCOCO/+/g \cite{kazemzadeh-etal-2014-referitgame,Mao_2016_CVPR}.}
\begin{tabularx}{0.9\linewidth}{
  l  
  |c  
  |*{3}{>{\centering\arraybackslash}X}  
  |*{3}{>{\centering\arraybackslash}X}  
  |*{2}{>{\centering\arraybackslash}X}  
}
\toprule
\multirow{2}{*}{\textbf{Method}} & \multirow{2}{*}{\textbf{Size}} & \multicolumn{3}{c|}{\textbf{RefCOCO}} & \multicolumn{3}{c|}{\textbf{RefCOCO+}} & \multicolumn{2}{c|}{\textbf{RefCOCOg}}  \\
\cmidrule(lr){3-5} \cmidrule(lr){6-8} \cmidrule(lr){9-10} 
&& val & testA & testB & val & testA & testB & val(U) & test(U)  \\
\midrule
\rowcolor{gray!10}\multicolumn{10}{l}{\textit{\textcolor{gray}{Non-LLM-based Specialists}}} \\
  \text{CoHD}$_{\text{[ICCV'25]}}$ \cite{Luo_2025_cohd} &-- & 78.11 &80.39 &75.20 &72.03& 76.37 &65.45 &70.83 &72.11  \\
\midrule

\rowcolor{gray!10}\multicolumn{10}{l}{\textit{\textcolor{gray}{LLM-based Image Generalists}}} \\
\text{GSVA}$_{\text{[CVPR'24]}}$\cite{xia2024gsva} & Vicuna-7B & 77.2 &78.9 &73.5& 65.9& 69.6& 59.8& 72.7& 73.3 \\
  
\text{SAM4MLLM}$_{\text{[ECCV'24]}}$\cite{Chen2025sam4mllm} & LLaVA-v1.6-8B & 79.8 & 82.7 & 74.7 & 74.6 & 80.0 & 67.2 & 75.5 & 76.4  \\

\text{SegAgent}$_{\text{[CVPR'25]}}$\cite{zhu2025segAgent} & LLaVA-v1.5-7B & 79.69 & 81.35 & 76.57 & 72.49 & 75.80 & 66.89 & 75.11 & 75.20  \\

\text{Text4Seg}$_{\text{[ICLR'25]}}$\cite{lan2025textseg} & InternVL2-8B & 79.2 & 81.7 & 75.6 & 72.8 & 77.9 & 66.5 & 74.0 & 75.3 \\
\midrule

\rowcolor{blue!7.5} \textbf{\model} (Ours) & Qwen2.5VL-3B & 78.86 &  80.23 &  75.15 & 73.09 & 77.38 & 66.98 & 74.06 & 75.08  \\

\rowcolor{blue!7.5} \textbf{\model} (Ours) & Qwen2.5VL-7B & 79.82 & 81.22 & 76.84 & 74.73 & 78.92 & 68.31 & 75.53 & 76.31  \\

\midrule

\rowcolor{blue!7.5} \textbf{\model} (Ours) & Qwen3-VL-4B & \underline{80.50} & \bf 82.75 & \bf 78.32 & \underline{75.13} & \underline{80.10} & \underline{69.23} & \underline{76.89} & \underline{77.97}  \\

\rowcolor{blue!7.5} \textbf{\model} (Ours) & Qwen3-VL-8B & \textbf{81.19} & 82.64 & \underline{78.07} & \textbf{75.88} & \textbf{80.24} & \textbf{70.01} & \textbf{77.23} & \textbf{78.04} \\
\bottomrule
\end{tabularx}
\label{tab:supply_results_base}
\end{table*}

To assess the generality of our \model framework across foundation model scales and architectures, we further evaluate its performance using multiple variants of the Qwen-VL family. In the main paper, we focused on the latest Qwen3-VL-based results due to their state-of-the-art performance. Here, we provide complementary results using Qwen2.5VL-based models in Table~\ref{tab:supply_results_base} to illustrate the behavior of our approach under earlier backbone settings.

\subsubsection{Performance on Qwen2.5VL-based \model}

With the lightweight Qwen2.5VL-3B backbone, our \model achieves competitive performance across all RES benchmarks, delivering results such as 78.86\% / 80.23\% / 75.15\% on RefCOCO (val/testA/testB), 73.09\% / 77.38\% / 66.98\% on RefCOCO+, and 74.06\% / 75.08\% on RefCOCOg (val(U)/test(U)). These results surpass several LLM-based generalist approaches (e.g., GSVA, Text4Seg) despite using a smaller backbone and no task-specific training.

Scaling to Qwen2.5VL-7B further improves performance, yielding up to 79.82\% (RefCOCO val) and 76.31\% (RefCOCOg test(U)), narrowing or surpassing the gap with concurrent models that rely on heavier backbones or auxiliary training (e.g., SegAgent). This highlights the strong compatibility of our method with mid-sized vision-language backbones.

\subsubsection{Enhanced Performance with Qwen3-VL-based \model}

Transitioning to Qwen3-VL as the base model brings consistent and significant gains. The Qwen3-VL-4B variant already outperforms all Qwen2.5VL-based experiments, achieving state-of-the-art performance on multiple splits (e.g., 82.75\% on RefCOCO testA and 78.32\% on testB).

The largest variant, Qwen3-VL-8B, further advances performance across the board, achieving 81.19\%/75.88\%/77.23\% on RefCOCO (val), RefCOCO+ (val), and RefCOCOg (val(U)), respectively. These results match or exceed the best reported performance in the literature, demonstrating that our framework can fully leverage advances in base model capacity and visual-text alignment.

\subsubsection{Summary and Insights}

These supplementary results validate that our method remains effective and competitive across base model versions and parameter scales. The performance gains observed when transitioning from Qwen2.5VL to Qwen3-VL are consistent with the expected improvements in visual-language understanding and spatial reasoning inherent to the newer foundations.

Importantly, our method does not rely on base model-specific modifications or task-dependent training, making it a lightweight and modular enhancement that can be seamlessly integrated with both existing and future multimodal large language models. This supports the central claim of our work: that geometry-aware point reasoning provides complementary capabilities to vision-language models, independent of their scale.

\section{Visualization}
\subsection{Successful Case Visualization and Mechanism Validation}
\label{sec:success_cases}

\begin{figure*}[t]
  \centering
   \includegraphics[width=\linewidth]{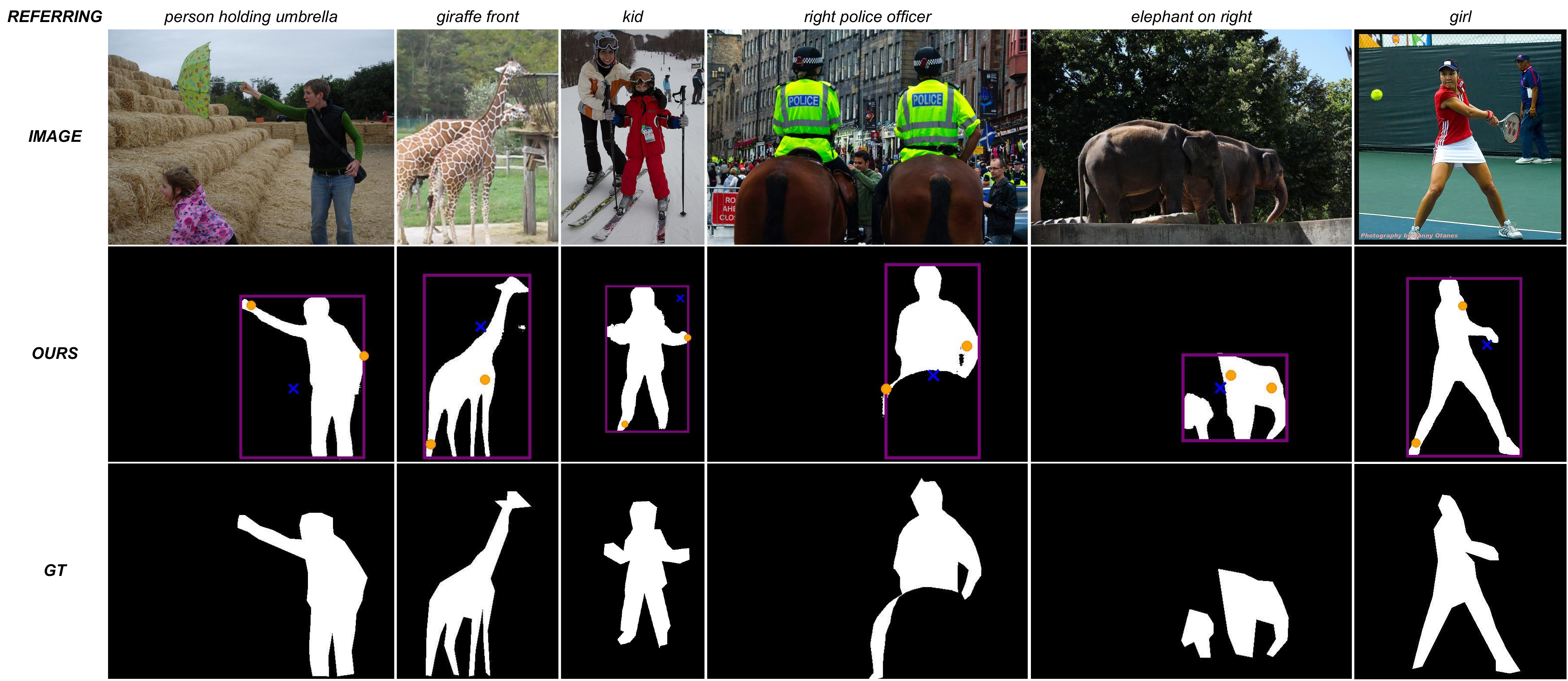}

   \caption{Successful case visualizations showcasing the Entropy-Based Point Discovery (EBD) strategy. For each example, rows display the referring expression, input image, our segmentation result, and GT. Yellow dots denote positive candidate points, purple box denotes the predicted bbox, and blue crosses denote negative candidate points.}
   \label{fig:supply_success_cases}
\end{figure*}

To further validate the mechanism of our Entropy-Based Point Discovery (EBD) strategy and illustrate its effectiveness in practice, we present a set of successful qualitative examples in Fig.~\ref{fig:supply_success_cases}. These visualizations cover diverse referring expression segmentation (RES) scenarios, including attribute grounding, spatial disambiguation, and background suppression. Together, they explicitly reflect how our entropy-guided reasoning produces high-value prompts for segmentation.

\subsubsection{Mechanism-Centric Case Analysis}
We highlight two representative examples that best demonstrate the core design principles behind EBD, including entropy-driven sampling, dual-queue discovery, and geometric uncertainty modeling.

\noindent\textbf{Case 1: Fine-Grained Edge Grounding (\emph{“person holding umbrella”})}\\
This example demands precise localization of fine structures (e.g., the umbrella handle and the person’s arm). Guided by the entropy proxy, the superellipse spiral efficiently samples boundary-proximal regions first, placing \textcolor{orange}{positive points} along the umbrella and arm contours. These points convey high-information membership evidence to the SAM module. Meanwhile, \textcolor{blue}{negative points} are detected in non-target interior regions (e.g., the gap between the arm and umbrella), constraining the candidate space and preventing mask leakage.

The interplay of edge-focused positive cues and carefully placed negative feedback enables SAM to produce a mask that nearly matches the GT. This example showcases how EBD’s structured uncertainty prior leads to effective fine-grained semantic grounding.

\noindent\textbf{Case 2: Spatial Disambiguation via Central Negative Discovery (\emph{“right police officer”})}\\
In this scene, the target (“right police officer”) overlaps with a foreground horse, resulting in spatial ambiguity. EBD addresses this by first allocating negative focus to the central region using the internal-queue strategy. Early detection of \textcolor{blue}{negative points} in the horse region prevents mis-segmentation and narrows the candidate scope. Follow-up \textcolor{orange}{positive points} along the right officer’s boundary enable fine localization without redundant attempts.

This case demonstrates how dual-queue design (external edge-first + internal center-first) effectively balances semantic precision and spatial exclusion, supporting robust disambiguation without detector-level supervision.

\vspace{1ex}
Across all examples, the segmentation outputs of our \model closely match the GT, validating that EBD’s entropy-maximizing sampling and geometry-aware reasoning are both practically effective and theoretically grounded. The visual evidence aligns with our findings from analytical studies and quantitative benchmarks.

\subsection{Failure Case Analysis}
\label{sec:failure_analysis}

\begin{figure*}[t]
  \centering
   \includegraphics[width=\linewidth]{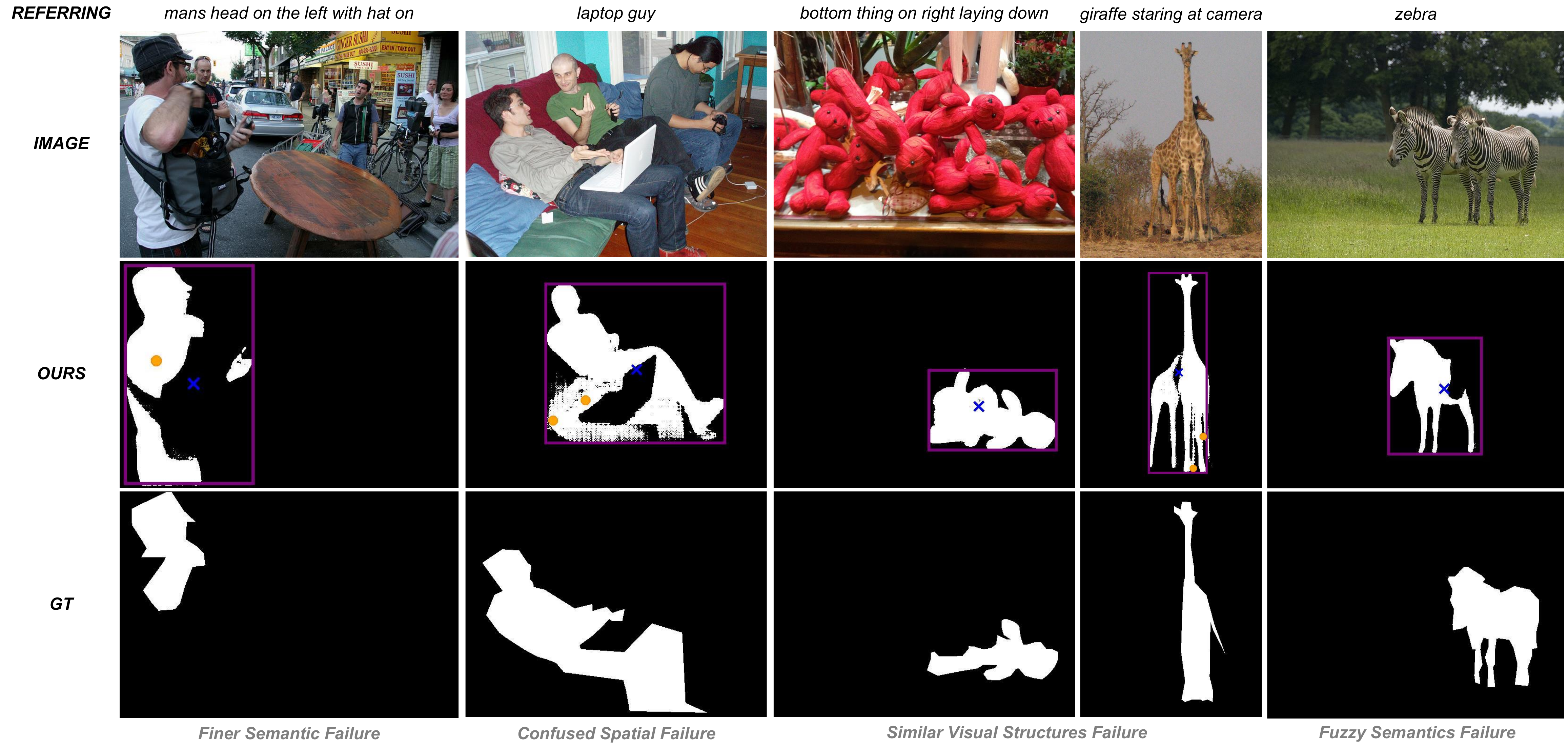}

   \caption{Failure case visualization of our \model. Examples are grouped by failure type (left to right), corresponding to (i) Finer Semantic Failure, (ii) Confused Spatial Failure, (iii) Similar Visual Structures Failure, (iv) Similar Visual Structures Failure, (v) Fuzzy Semantics Failure. Each example shows the original image, textual-based segmentation, VBR prediction, and ground-truth mask (top-to-bottom).}
   \label{fig:supply_failure_cases}
\end{figure*}

To further investigate the limitations of our proposed visual behavior reasoning (VBR) approach, we visualize several representative failure cases in Fig.~\ref{fig:supply_failure_cases}, categorized into four primary patterns. Each example highlights a distinct failure mode in resolving referring expressions, revealing both algorithmic limitations and inherent challenges within RES datasets.

\paragraph{Finer Semantic Failure.}
In this example, the expression ``mans head on the left with hat on'' requires precise understanding of fine-grained spatial semantics (e.g., \emph{“head”} vs.\ \emph{“whole person”}). While our model correctly localizes the target person, it over-segments the entire figure rather than isolating the head region. This suggests a limitation in extracting highly localized semantic attributes, especially when the target is embedded within a larger entity.

\paragraph{Confused Spatial Failure.}
The expression ``laptop guy'' is ambiguous due to multiple individuals simultaneously interacting with laptop-like objects. Our model segments the wrong person due to ambiguous spatial cues and the lack of sufficiently discriminatory language information. Even the ground truth only includes the lower half of the target person, highlighting inconsistencies in annotation and implicit subjective decisions.

\paragraph{Similar Visual Structures Failure.}
In the third and fourth cases, similar visual objects are densely packed (e.g., multiple red plush toys or overlapping giraffes). The model struggles to resolve the exact target instance due to near-identical appearance and insufficient discriminative references in the language. These failures are not strictly due to algorithmic errors but result from lack of unique visual or linguistic cues.

\paragraph{Fuzzy Semantics Failure.}
The final case shows inherent ambiguity in the referring text: the expression ``zebra'' refers to a scene with two equally salient zebras. Our model segments the left zebra, while the ground truth labels the right one. This discrepancy underscores a limitation of RES datasets—when expressions are under-specified, even perfect visual reasoning cannot resolve ambiguity without richer linguistic grounding or contextual priors.

Overall, these failure cases expose a spectrum of challenges in RES: from fine-grained attribute grounding and precise spatial disambiguation to visual similarity and textual vagueness. Crucially, several failures stem from inherent ambiguity or noise in the data itself, suggesting that future progress will require not only better models but also improved annotation standards and richer language supervision.

\section{Implementation Details}
\label{sec:imple_details}

\subsection{MLLM Details}

\subsubsection{Model Selection and Usage}
We adopt the Qwen-VL family as the backbone multimodal language model, owing to its strong visual-language alignment and accurate spatial grounding. Specifically, we experiment with four representative variants: Qwen2.5-VL (3B / 7B) and Qwen3-VL (4B / 8B).

Crucially, the MLLM is \emph{not fine-tuned end-to-end}. Instead, we introduce a lightweight Visual Binding Reasoning (VBR) module trained solely in a VQA paradigm to enhance point-wise visual grounding. All bounding box (bbox) predictions are handled by the pre-trained Qwen-VL models, without any further bbox-specific training. This preserves the original model's pretrained spatial reasoning capability while minimizing tuning overhead.

\subsubsection{Dataset Construction for VBR Training}
To support VBR via binary visual QA, we construct a training dataset based on the RefCOCO, RefCOCO+, and RefCOCOg train splits. We sample 15\% of dataset instances from each split and generate multiple candidate point queries per instance.

For each sampled image, internal and external candidate points are generated using the EBD strategy, averaging twice the number used during inference to encourage diversity. Each point is visually marked using a randomly selected color (from 8 distinct options) and shape (star, circle, or hexagon), with marker size randomly drawn from the interval $[6,24]$ px. To further improve generalization, 50\% of marked images are randomly cropped around the marker. This facilitates variable resolution and context-robust reasoning.

Each marker is paired with a binary VQA prompt:
\begin{quote}
\small
\emph{``Answer strictly yes or no: Is the \{\textit{color}\}-colored \{\textit{marker}\} on the object referred to by `\{$\mathcal{T}$\}' in the picture?''}
\end{quote}
where ground-truth labels are assigned automatically based on the segmentation mask (inside = yes, outside = no). This dataset directly supports the learning of visual-textual point bindings under diverse spatial and semantic configurations.

\subsubsection{LoRA Fine-Tuning for VBR}
We adopt Low-Rank Adaptation (LoRA) for efficient fine-tuning, with a unified configuration across all Qwen-VL variants. Hyperparameters are summarized in Table \ref{tab:lora_params}.

\begin{table}[h]
\centering
\small
\begin{tabular}{lc}
\hline
\textbf{Parameter} & \textbf{Value} \\
\hline
LoRA Rank & 8 \\
Target Layers & All \\
Per-Device Batch Size & 1 \\
Gradient Accumulation & 8 \\
Learning Rate & $1.0\times10^{-4}$ \\
Epochs & 1.0 \\
LR Scheduler & Cosine decay \\
Warmup Ratio & 0.1 \\
Training Samples & 80,000 (randomly sampled) \\
\hline
\end{tabular}
\caption{Unified LoRA fine-tuning hyperparameters for VBR training.}
\label{tab:lora_params}
\end{table}

To simulate a larger batch size and ensure balanced gradient updates, we use gradient accumulation (effective batch size = 8). Cosine LR decay with 10\% warmup enables stable convergence over the single epoch of training. Model checkpoints and logs are periodically saved during training to monitor progress.

\subsection{Decoder Details}

\subsubsection{Model Configuration}
We adopt SAM2 as the segmentation decoder, using the \texttt{Base\_Plus} and \texttt{Large} variants for all experiments. These variants offer an effective trade-off between mask quality and inference efficiency, particularly suitable for instance segmentation settings. The SAM2 model is trained independently and plugged into our pipeline without joint optimization with the MLLM or VBR modules, enabling efficient modular integration.

\subsubsection{COCO-to-SA-1B Format Conversion}
To train SAM2 on COCO instance masks, the COCO \texttt{train2014} split is converted into SA-1B format. The SA-1B format requires a one-to-one correspondence between images and annotation files. We process the COCO \texttt{instances\_train2014.json} file and prepare a separate annotation JSON per image. Each JSON contains two fields: \texttt{image} (metadata) and \texttt{annotations} (instance list).

\noindent\textbf{Step 1: Data Extraction \& Grouping.}
We extract image metadata (ID, size, filename) and instance annotations (masks, bounding boxes, area). Using \texttt{image\_id}, instance annotations are grouped per image.

\noindent\textbf{Step 2: Annotation Conversion.}
Each COCO instance annotation is converted to SA-1B format:

\begin{itemize}
    \item \textbf{Segmentation:} Convert COCO-format polygons or RLE masks to SA-1B RLE: \texttt{size}: $[H, W]$, \texttt{counts}: encoded as UTF-8.
    \item \textbf{Crop box:} Set to full image extent: \texttt{[0, 0, W, H]}.
    \item \textbf{Point coords:} Sample a non-zero mask pixel and record its $(x, y)$ coordinate (note COCO uses $(y, x)$ order).
    \item \textbf{Stability score:} Set to $1.0$, matching SA-1B's default high-confidence label.
\end{itemize}

\noindent\textbf{Step 3: Output Organization.}
For each image, a JSON file with the same base name is generated (e.g., \texttt{000000000009.jpg} and \texttt{000000000009.json}). This creates a fully SA-1B-compliant dataset consumable by SAM2 without modification.

\subsubsection{Training Setup}
SAM2 is trained following its default pipeline, with hyperparameters summarized in Table \ref{tab:training_params}. Mask prediction is supervised using a multi-task loss on binary masks, dice, IoU, and instance classification.

\begin{table}[h]
\centering
\small
\begin{tabular}{lc}
\hline
\textbf{Parameter} & \textbf{Value} \\
\hline
Input Resolution                          & $1024\times1024$ \\
Batch Size                                & $4$ \\
Worker Threads                            & $10$ \\
Frames per Sample                         & $8$ \\
Max Instances per Sample                  & $3$ \\
Base Learning Rate                        & $5.0\times10^{-6}$ \\
Vision Encoder LR                         & $3.0\times10^{-6}$ \\
Total Epochs                              & $40$ \\
Optimizer                                 & AdamW \\
Gradient Clipping                         & $0.1$ \\
Mixed Precision                           & bfloat16 (enabled) \\
Loss Weights (Mask:Dice:IoU:Class)        & $20:1:1:1$ \\
\hline
\end{tabular}
\caption{SAM2 training hyperparameters.}
\label{tab:training_params}
\end{table}

All other components (e.g., data augmentation, embeddings, early stopping) follow the default SAM2 configuration. This preserves compatibility with the official SAM2 training pipeline and ensures fair comparison to related works.

\section{Limitations and Future Directions}
\label{sec:limitations}

Despite the strong performance and generality of \model, we acknowledge several limitations that also highlight promising directions for future work.

\paragraph{Dependence on Bounding Box Priors.}
Our point discovery relies on an initial bounding box to constrain the search space. While this design enables task efficiency and we have demonstrated robustness under practical perturbations (Sec.~\ref{sec:supply_oracle_bbox_prior}), highly inaccurate box priors may still undermine spatial reasoning. Incorporating a lightweight box-refinement mechanism or exploring box-free point discovery could further enhance resilience in challenging settings.

\paragraph{Handcrafted Geometric Inductive Bias.}
The entropy-based sampling strategy is driven by a structured geometric heuristic (superellipse spiral and dual-queue search). Though validated across diverse benchmarks, its static form may not fully adapt to irregular or small-scale objects. Future extensions may include learning adaptive sampling patterns or integrating entropy estimation into the model itself.

\paragraph{No End-to-End Optimization Across Components.}
Finally, our framework integrates separately trained modules (VBR and decoder) without joint optimization. While advantageous for modularity, this also limits potential synergy between geometric point reasoning and visual mask generation. Future work could investigate lightweight coupling mechanisms to align point discovery and segmentation behavior while retaining the modular design benefits.

Overall, these limitations suggest promising directions for advancing \model toward more generalizable, adaptive, and interactive RES—from improved grounding without initial boxes, to entropic point discovery beyond geometric heuristics, and tighter coupling with downstream segmentation modules.

\section{Potential Impact and Future Research Directions}

Beyond its immediate performance gains in referring expression segmentation, \model introduces several concepts that may inform broader developments in multimodal perception and vision-language reasoning.

\paragraph{Entropy-Driven Interaction as a General Interface.}
Our formulation advocates an information-theoretic view of spatial interaction: rather than directly predicting segmentation masks or per-pixel labels, the model discovers informative visual cues by reasoning over an uncertainty field. This paradigm shifts the focus from dense prediction to targeted information extraction, and could serve as a foundation for lightweight, interactive segmentation in open-world applications, such as annotation-efficient dataset creation, human-in-the-loop editing, or embodied scene understanding.

\paragraph{Agent-Based Reasoning in Pixel Space.}
The use of an explicit agent to perform spatial selection and verification over pixel coordinates offers a new way of bridging MLLM reasoning with low-level vision tasks. Unlike token-based spatial grounding, which relies on indirect symbol manipulation, our approach affords interpretable, geometry-aware actions that could be extended to tasks such as object-level exploration, visual affordance detection, or interactive scene querying.

\paragraph{Modular Vision-Language Segmentation Framework.}
The modular design of \model—separating point discovery, visual verification, and segmentation—offers a reusable interface that can interoperate with future advancements in any single component, such as improved bounding box proposals, adaptive point strategies, or segmentation backbones beyond SAM. This flexibility suggests a practical path toward scalable, plug-and-play RES systems with minimal retraining overhead.

\paragraph{Broader Implications.}
The proposed agent-driven entropy maximization also hints at a more general strategy for resource-efficient vision-language interaction: allow the language model to drive where to look, and let the vision backbone refine what to segment. We believe this direction may inspire future work toward compositional, action-conditioned models that better integrate reasoning and perception, particularly in embodied or long-term interactive environments.

Overall, \model demonstrates that integrating uncertainty-aware geometric reasoning with foundation models is not only effective for RES but also carries broader implications for multimodal active perception and interpretable pixel-level decision-making.


\end{document}